\documentclass[lettersize, journal]{IEEEtran}                                                          

\IEEEoverridecommandlockouts                              
\usepackage{caption}
\usepackage{subcaption}
\usepackage{color}
\usepackage{amsmath,amssymb}
\usepackage{todonotes}
\usepackage{algorithm, algorithmic}
\usepackage{tabu}
\usepackage{booktabs}
\usepackage{graphicx}
\usepackage{bbm}
\usepackage{cite}
\usepackage{dirtytalk}
\usepackage{ulem}
\usepackage{hyperref}

\newcommand{\appropto}{\mathrel{\vcenter{
  \offinterlineskip\halign{\hfil$##$\cr
    \propto\cr\noalign{\kern2pt}\sim\cr\noalign{\kern-2pt}}}}}
\newcommand{\realspace}{\mathbb{R}}

\newcommand{\TROEdits}[1]{{\color{black} #1}}

\title{HARPS: An Online POMDP Framework for Human-Assisted Robotic Planning and Sensing}

\author{Luke Burks, Hunter M. Ray, Jamison McGinley, Sousheel Vunnam, and Nisar Ahmed$^*$
\thanks{$^*$Smead Aerospace Engineering Sciences Department, 429 UCB, 3775 Discovery Drive, University of Colorado Boulder, Boulder CO 80309, USA. E-mail:{\ttfamily[luke.burks;nisar.ahmed]@colorado.edu}. This work is funded by the Center for Unmanned Aircraft Systems (C-UAS), a National Science Foundation Industry/University Cooperative Research Center (I/UCRC) under NSF Award No. CNS-1650468 along with significant contributions from C-UAS industry members.
 }
}

\begin{document}


\maketitle

\begin{abstract}
The ability of autonomous robots to model, communicate, and act on semantic ‘soft data’ remains challenging. The Human-Assisted Robotic Planning and Sensing (HARPS) framework is presented for active semantic sensing and planning in human-robot teams to address these gaps by formally combining the benefits of online sampling-based POMDP policies, multimodal human-robot interaction, and Bayesian data fusion. HARPS lets humans impose model structure and extend the range of soft data by sketching and labeling new semantic features in uncertain environments. Dynamic model updating lets robotic agents actively query humans for novel and relevant semantic data, thereby improving model and state beliefs for improved online planning. Simulations of a UAV-enabled target search in a large-scale partially structured environment show significant improvements in time and beliefs required for interception versus conventional planning with robot-only sensing. A human subject study in the same environment shows an average doubling in dynamic target capture rate compared to the lone robot case, and highlights the robustness of HARPS over a range of user characteristics and interaction modalities. 
\end{abstract}


\section{Introduction}

Autonomous robotic vehicles will greatly extend human capabilities in domains such as space exploration \cite{mcguire2018failure}, disaster response \cite{delmerico2019current}, environmental monitoring \cite{dunbabin2012robots}, infrastructure inspection \cite{hutter2018towards}, search and rescue \cite{GoodrichSAR_UAS}, and defense \cite{david2016defense}. Yet, the uncertain and dynamic nature of these settings, coupled with vehicle size, weight, power, and compute constraints (SWaP-C), often necessitates some form of human oversight to cope with the brittleness of autonomy \cite{bradshaw2013seven}. This has created interest in new forms of human-robot interaction that can efficiently leverage human input to enhance robotic reasoning abilities. Probabilistic techniques relying on semantic language-based human-robot communication have gained considerable attention for information fusion \cite{bourgault2008scalable,kaupp2010human,rosenthal2011learning,bishop2013fusion,ahmed2013bayesian, sweet2016structured} and task planning \cite{skubic2003sketch,tellex2011understanding,shah2012sketch, howard2014natural, duvallet2016inferring, paul2016efficient}. 
However, existing approaches 
only allow robots to reason about limited kinds of uncertainties, i.e. as long as task environments and conditions are known a priori, or do not change in unforeseen ways. These approaches are antithetical to how human teams communicate with subtle ambiguities to express uncertainty. Constraining the types of information limits the flexibility and utility of semantic communication for adapting to new or unknown situations.

This work examines how robots operating in uncertain environments can use semantic communication with humans to solve the combined issues of model augmentation, multi-modal information fusion, and replanning under uncertainty in an online manner. 
Such problems practically arise, for instance, with time-sensitive dynamic target search in areas with outdated or poorly defined maps, which lead to uncertain robot and target motion models as well as uncertain human input models. 
 A novel framework, Human-Assisted Robotic Planning and Sensing (HARPS), is presented for integrated active semantic sensing and planning in human-robot teams that formally combines aspects of online partially observable Markov decision process (POMDP) planning, sketch and language-based semantic human-robot interfaces, and model-based Bayesian data fusion. 
 
\begin{figure*}[t]
	\centering	
	\begin{subfigure}{.49\textwidth}
		\includegraphics[width=\textwidth]{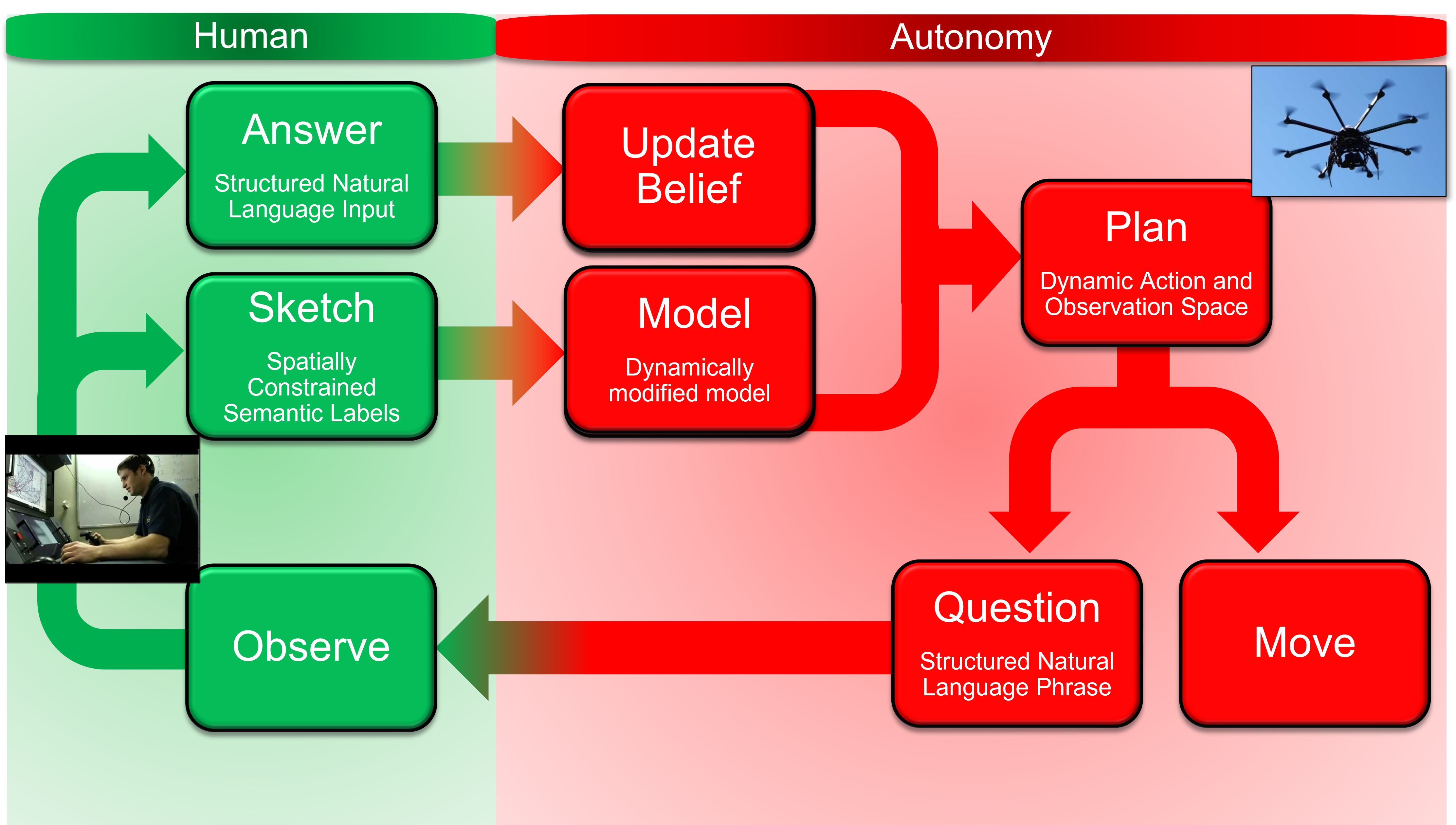}
    \end{subfigure}
    ~
    \begin{subfigure}{.49\textwidth}
    	\includegraphics[width=\textwidth]{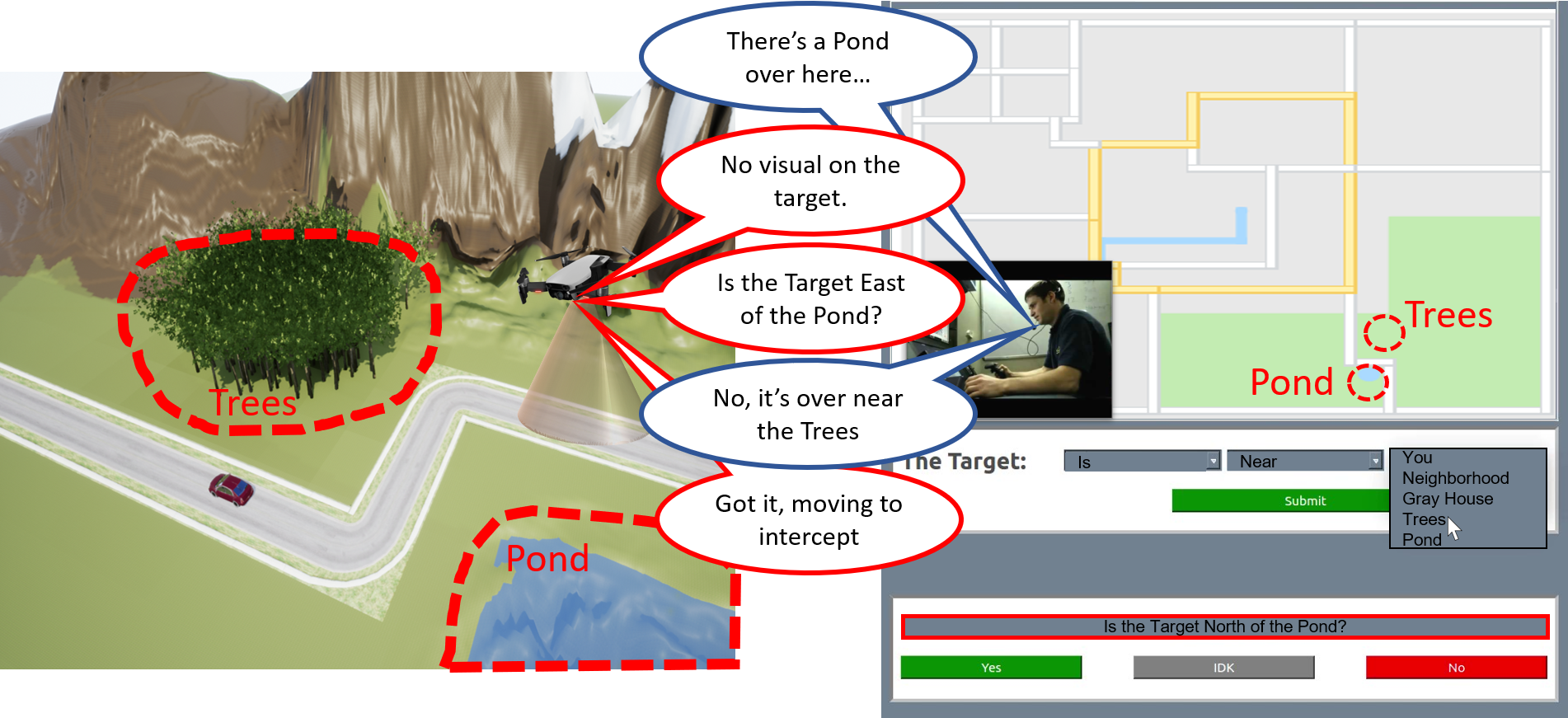}
    \end{subfigure}
    \caption{A scenario showcasing the novel contributions of the HARPS framework including sketching, querying, planning, and modeling in the context of dynamic dictionaries and environments. A video of the human interface and simulation environment can be found at:  \url{https://youtu.be/kaQnoM3GTXo}.}
    \label{fig:marquee}
\end{figure*}

As shown in Fig. \ref{fig:marquee}, HARPS features three key technical innovations which are developed and demonstrated in the context of dynamic target search tasks in uncertain environments. 
Firstly, humans act as `ad hoc' sensors that push multi-\TROEdits{modal} semantic data to robots via structured language, enabling robots to update task models and beliefs with information outside their nominal sensing range.  
This information can take the form of state observations, target dynamics, and problem structure, which can be represented \TROEdits{by different modes in} both robotic and human reasoning. Such distinctions allow multiple modes of information to be communicated and fused in an intuitive fashion while maximizing their utility to the robot. 
Secondly, humans can use real-time sketch interfaces to update semantic language dictionaries grounded in uncertain environments, thus dynamically extending the range and flexibility of their observations. 
Finally, robots actively query humans for specific semantic data on multiple modes to improve online performance, while also  planning \TROEdits{with foresight} to act with imperfect human sensors. 
These features effectively enable online `reprogramming' of POMDPs together with human-robot sensor fusion to support online replanning in complex, dynamic environments. 
%
%

This work introduces, describes, and demonstrates \TROEdits{three} novel technical contributions to the state of the art: 
\begin{enumerate}
    \item A novel framework for multi-modal human-robot collaboration through structured semantic language interfaces;
    \item \TROEdits{Dynamic, sketch-based modification of an online POMDP, evolving previous work \cite{burks2019collaborative} with novel computation allocation};
    \item Validated approach through human-subject studies on a simulated dynamic target search problem.
\end{enumerate}

These contributions significantly extend the earlier work in \cite{burks2019collaborative} in several important respects. 
Firstly, although \cite{burks2019collaborative} introduced the concept of sketch-based modification of online POMDPs, it only considered limited types of dynamic semantic observation dictionaries that can be provided and fused by autonomous robots. The HARPS framework presented here not only accounts for collaborative semantic data fusion pertaining to dynamic state variables but also dynamic model and environment variables that impact robot motion and higher level target state behavior in search problems. 
Secondly, while \cite{burks2019collaborative} demonstrated a limited set of results in a simplified 2D simulated environment, this paper presents a more general and scalable implementation that is amenable to resource-constrained robots. This is also demonstrated here in a larger physics-based 3D simulation, using more realistic robot motion, environmental, and sensing models. 
Finally, whereas our earlier work focused only on the algorithmic aspects of online modeling, data fusion, and planning, this paper also applies HARPS in a thorough set of synthetic and live human subject studies to validate our approach and gain better insight into its utility for practical robotics problems.

This paper is organized as follows: Sec. II presents motivating dynamic search problems and reviews methods for planning and semantic data fusion under uncertainty; Sec. III details the formal problem statement; Sec. IV describes technical details of the HARPS framework; Sec. V presents target search simulation results with simulated human sketch and structured language observations; Sec VI presents the human subject study; and Sec. VII presents conclusions.

\section{Motivating Problems and Background}
%
%

\TROEdits{This work contributes an algorithmic framework that can be generalized to a variety of human-robot teaming scenarios. As robots gain greater autonomous capabilities, they need to recognize how to engage with their human collaborators to extract information about the environment and task at hand. Whether the operator acts in a supervisory or collaborative capacity, our framework enables their observations to augment autonomous robotic decision making. Facilitating robotic autonomy frees operators to focus on a variety of tasks, including allowing for a one-to-many framing of human-robot teaming \cite{lewis2009using}, or coordinating operation within the context of a larger incident or team, such as in public safety\cite{ray2022UASPublicSafety}.

During an evolving collaborative task between a human and robot, natural uncertainties may arise due to incomplete prior information, including object location and associated dynamics models. Resolving these inconsistencies through semantic language provides an intuitive method of collaboration. However, robots need to be flexible with understanding new reference locations that may not have been known at the start, and incorporate this variable and inconsistent human-generated `soft data' into their planning \cite{rosenthal2011learning},\cite{ sample2012experimental}}.

\begin{figure}[t]
\centering	
    \includegraphics[width=.3\textwidth]{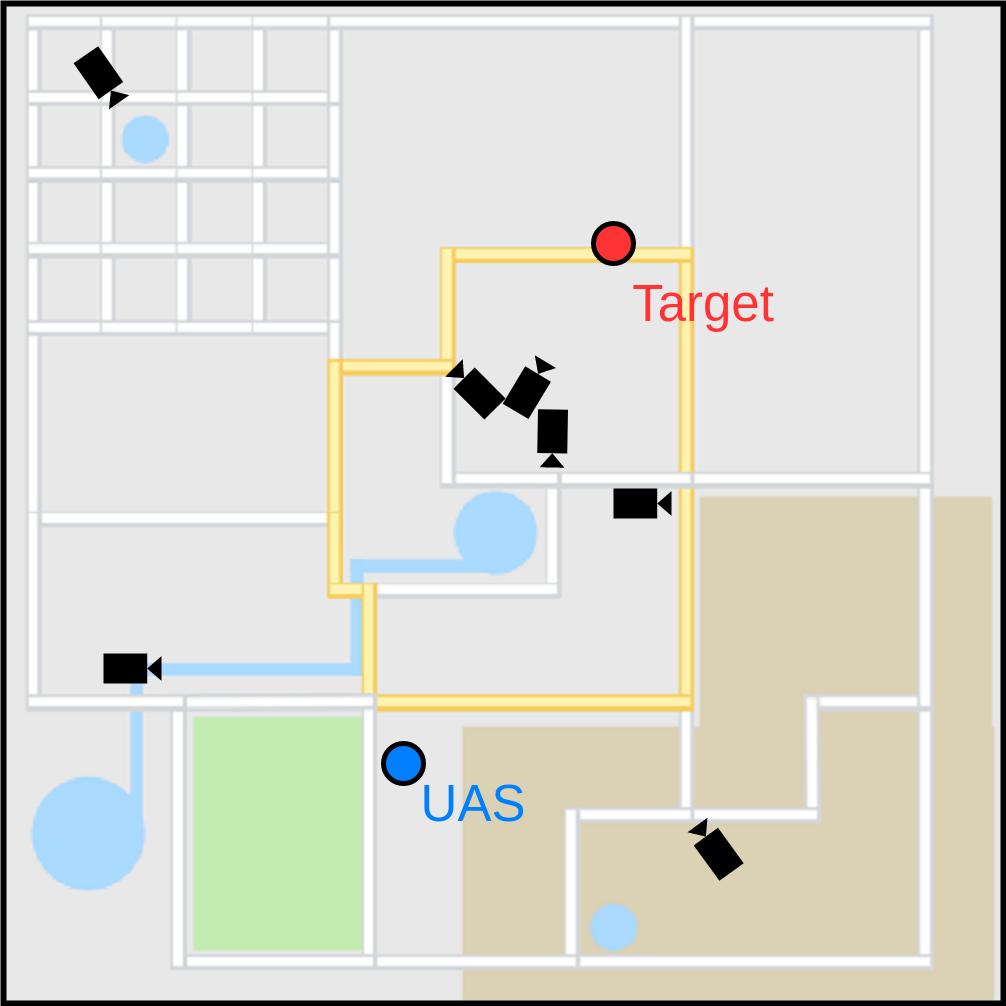}
	\caption{\TROEdits{Our Problem Setup incorporates a UAS tracking} a mobile ground target that can deviate from road. The user has access to live camera views across the environment, which they use to provide relevant information to help the UAS improve its search.}
    \label{fig:bothAnnotated}
\end{figure}

\TROEdits{This generalized problem is applied to a specific application of a dynamic target search task in which a uncrewed aerial system (UAS) must autonomously intercept a moving target as quickly as possible.} The UAS must attempt to localize, track, and intercept a ground based target within the span of its limited flight time. The environment\TROEdits{, shown in Figure \ref{fig:bothAnnotated},} consists of a large, open space with a variety of distinct regions such as a farm, a neighborhood, mountains, hills, plains, and rivers. The robot only has a high level map of the area available at the beginning of the search, such as a commercially available digital atlas, rendering specific locations and spatial extents of structures or landmarks within the environment unknown or incomplete a priori. The target can either drive through the local road network, which is fast but more spatially constrained, or proceed on foot in a slower yet unconstrained fashion. \TROEdits{A remote operator is available to help locate the target, and, with the help of various surveillance cameras, can communicate semantic information to the UAS regarding the target's location, either upon request or in a volunteer capacity. They can also view the aircraft's telemetry and video feed. 

The UAS carries an optical sensor, which allows for high probability target proximity detection that is subject to false alarms. It uses this information in conjunction with the operator's semantic observations to update its belief of the target location before planning its own movements.}

\subsection{Language-based and Sketch-based Semantic Soft Data}
The human primarily acts as an auxiliary and imperfect semantic information source that can interface with the robot at any time via either one of two methods. 
The first interface allows the human to compose linguistic statements that are parsed and interpreted as target state observations. As shown in Fig. \ref{fig:marquee}, these are modeled in the structured form `Target \emph{is/is not} \emph{Desc} \emph{Ref}', where the \emph{Desc} and \emph{Ref} elements are taken from a defined \TROEdits{set of semantic terms, i.e a codebook,} and the \emph{is/is not} field toggles between positive and negative data \cite{koch2007exploiting}. In prior work \cite{sample2012experimental,ahmed2013bayesian,sweet2016structured,lore2016deep,burks2018closed}, probabilistic likelihoods were developed for all possible statements in a given codebook to support recursive Bayesian fusion of linguistic human and robot sensor data. Since these works considered only \textit{fully known} environments, all relevant semantic references could be enumerated in advance. However, in the motivating problem the space has not been mapped in high detail prior to the search, so the codebook is only partially defined at mission start. While this initially reduces the information available to the robot, the capability to reason and learn with sparsely detailed maps provides new opportunities for practical implementation.

This leads to the second interface. In cases where the recognition of key environmental features are not readily captured by robotic perception, due to factors such as on-board compute constraints or limited training data, codebook augmentations can instead be obtained from a human providing labeled 2D free-form sketches, which each depict a spatially constrained region on a 2D map display (as in Fig. \ref{fig:marquee} with `Pond' and `Trees'). 
Building on \cite{burks2019collaborative}, the codebook is automatically augmented with the labels of new landmarks, so that the corresponding spatial sketch data can also be used to generate suitable soft data likelihood models for the linguistic statement interface (see Fig. \ref{fig:marquee}). However, unlike in \cite{burks2019collaborative}, human sketches here may also provide direct information about probabilistic state transition models to constrain how the robot and target may traverse certain map areas. While similar 2D sketch interfaces have also been developed for robotics applications \cite{skubic2003sketch,boniardi2015robot,shah2012sketch,ahmed2015fully}, 
this work presents a novel use case in the context of multi-level data fusion for planning under uncertainty. 

\subsection{Optimal Planning and Sensing under Uncertainty}
The robot must autonomously generate its own plans for minimum time target interception, whether or not any semantic soft data are provided. 
The presence of model uncertainties, sensing errors, and process noise makes optimal planning quite challenging. 
One family of decision-making algorithms that accounts for these combined uncertainties are partially observable Markov decision processes (POMDPs). 
While POMDPs are impractical to solve exactly for all but the most simple problems \cite{kaelbling1998planning}, a variety of powerful approximations 
can exploit various features of particular problem formulations. 
%
A POMDP is formally specified as a 7-tuple $(\mathcal{S,A,T,R},\Omega,\mathcal{O,\gamma})$, where the goal is to find a policy $\pi$ which maps from a Bayesian posterior distribution, i.e. the belief $b=p(s)$ over the set of states $\mathcal{S}$ with probability distribution $p$, to a discrete action $a \in \mathcal{A}$. 
The transition model $\mathcal{T}$ is a discrete time probabilistic mapping from one state to the next given an action, $p(s'|s,a)$, after which the robot is rewarded according to $R(s,a)$. During policy execution, the robot receives observations $o \in \Omega$ according to observation likelihood $\mathcal{O} = p(o|s)$. 
For infinite horizon planning problems with discount factor $\gamma \in [0,1)$, and discrete time step $k \in \mathbb{Z}^{0+}$ 
the optimal policy $\pi[b(s)] \rightarrow a$ maximizes the 
expected future discounted reward: 
$\mathbb{E}[\sum_{k=0}^{\infty} \gamma^{k} R(s_{k},a_{k})]$.

The primary challenge arising from casting our motivating problem as a POMDP lies in the ability of the human to modify the models $\mathcal{T}$ and $\mathcal{O}$ online in an unmodeled ad hoc fashion via the sketch interface. These modifications can happen rapidly, and might change large swathes of each model with a single sketch. Bayes-Adaptive POMDPs \cite{ross2008bayes} allow POMDPs to learn the parameters of $\mathcal{T}$ and $\mathcal{O}$, but require gradual changes to their parameters and furthermore assume static $\mathcal{T}$ and $\mathcal{O}$. In the motivating problem,
$\mathcal{A}$, $\mathcal{O}$, and $\mathcal{T}$ may change unpredictably with new sketches. 
This issue 
renders 
``full-width" offline point-based POMDP planners \cite{pineau2003point,spaan2005perseus} inapplicable, as they require models of how $\mathcal{T,O}$ and $\Omega$ change. 

Online POMDP approaches \cite{silver2010monte}, which eschew the process of pre-solving the policy prior to execution in favor of interleaving steps of policy execution and search, have successfully been used to address large observation spaces \cite{sunberg2018online}, continuous state spaces \cite{goldhoorn2014continuous}, and more recently, dynamic ad hoc models \cite{burks2019collaborative}. These algorithms make use of a `black-box' generative model, requiring only the `current POMDP problem' at time of execution, making them good candidates for solving problems with dynamic model uncertainty.

\subsection{Active Semantic Sensing for Planning with Human Collaborators}
A major challenge for problems like target localization is that dynamics and uncertainties can easily become quite non-linear and non-Gaussian, particularly given the types of semantic information available for fusion (e.g. negative information from `no detection' readings \cite{koch2004negative}). 
As a result, typical stovepiped approaches to control/planning and sensing/estimation can lead to poor performance, since they rely on overly simplistic uncertainty assumptions. 
Constraints on human and robot performance also place premiums on when and how often collaborative data fusion can occur. 
For example, it is generally important to balance situational awareness and mental workload for a human sensor (who might also need to switch between tasks constantly). Likewise, it is important for the robot to know how and when a human sensor can be exploited for solving complex planning problems, which would otherwise be very inefficient to tackle using only its own local sensor data. POMDP formulations offer a rigourous pathway to simultaneously addressing these issues, though approximations must still be considered for practical implementation \cite{burks2021collaborative}.

One approach to POMDP approximation, the QMDP algorithm \cite{littman1995learning}, attempts to use a fully observable MDP policy to compute the optimal action in a partially observed step. As QMDPs are only exactly optimal assuming the state will indeed become fully observable after a single timestep, they are generally unsuitable for information gathering dependent problems such as the one addressed in this work. However, the introduction of the oracular POMDP (OPOMDP) formulation \cite{armstrong2007oracular,armstrong2008approximate} builds on a QMDP policy and enables the use of a human sensor to provide `perfect' state information at a fixed cost. Further work resulted in Human Observation Provider POMDPs (HOP-POMDPs) \cite{rosenthal2011modeling}, which allow the consideration of oracular humans who are not always available to answer a robotic query. HOP-POMDPs calculate a cost of asking, which is then weighed against the potential information value, similar to VOI aware planning in \cite{kaupp2010human}. When augmented with the Learning the Model of Humans as Observation Providers (LM-HOP) algorithm \cite{rosenthal2011learning}, HOP-POMDPs can estimate both the accuracy and availability of humans, thus treating them as probabilistic sensors. A primary drawback to using either OPOMDPs or HOP-POMDPs to address target tracking problems is that while they both enable a QMDP based policy to consider information gathering actions, these actions consist of a single self-localization query. Such formulations ignore the rich information set available in the motivating problem thanks to the presence of semantically labeled objects.  The present work addresses this limitation by expanding the information gathering actions to include a broader set of queries regarding target position and dynamics. 

\subsection{Planning for Bi-directional Communication}

Human-robot communication presents unique challenges in reasoning with respect to the complex nuances of natural language and human interaction. To perform comparably to an all-human team, robots working with humans benefit greatly from the ability to actively query their teammates to reason about given commands, aid collaboration, and update their understanding about the task or environment. There is a substantial body of work on different aspects of natural language processing, for which \cite{Tellex2020RobotsTU} provides a comprehensive review. Our motivating problem is similar to that of language grounding, where a system needs to reference a given command to a location within the environment. Methods proposed in \cite{tellex2011understanding} and \cite{howard2014natural} implement graphical models to reason over a teammate's commands to aid in task planning. Spatial distribution clauses are used to infer a location based hierarchy that allows a robot to reason over a wide range of commands. However, these approaches are limited by a static vocabulary that doesn't allow for the entry of new spatial or semantic information into the environment. As an environment may change over the course of an interaction or new environments are encountered, the vocabulary used to describe such environment must also be able to adapt, which is a focus of more recent work such as \cite{Berg2020GroundingLT}. We obviate some of the challenging components of natural language processing by having the operator define physical locations using sketches in real-time on a map, which the robot then uses as a spatial reference for integrating the provided observations and reasoning over queries to the operator.

While predefined interfaces, such as GPS coordinates, could be used to reason in a precise manner about geographic locations, this type of deterministic communication is unintuitive and prone to errors. Human teams will often communicate with inherent ambiguity necessitating follow up questions between teammates to ensure comprehension. However, reasoning over this type of nuanced information remains challenging for robots. The method devised in \cite{tellex2011understanding} set up a framework for comprehending \textit{what} information the \textit{user} provided, whereas \cite{Unhelkar2019HRC},\cite{Unhelkar2020BidirectionalCommunication} and \cite{Nikolaidis2018PlanningwithVerbal} consider the problem of \textit{if, when,} and \textit{what} to communicate \textit{back} to the user in order to resolve ambiguity across an interaction. These approaches utilize POMDPs to reason about some component of the human's mental state in different collaborative tasks. The work presented in \cite{rosen2020MixedReality} shows how a POMDP can be utilized to refine the robot's understanding of a human's object of interest. In \cite{Arkin2018RealTimeHC} a robot attempts to infer partially observable states of certain objects. However, in each of these cases the user is limited in their vocabulary. These methods constrain the robot to communicate about a static, and potentially obsolete environment, especially when applied in a dynamic target search where new locations become relevant to planning or querying.

In uncertain environments, new spatial information may be presented that, when reasoned over, can provide critical information to the completion of a task.  The approach described in this work uses a simplified compass based spatial reference cues (North, South, Near, etc.) coupled with a dynamic vocabulary codebook that the user can update and modify in real time, allowing the user to communicate to the robot teammate in an intuitive manner. Furthermore, the use of a POMDP to select the nature and timing of human communication enables reasoning with respect to the Value of Information theory proposed in \cite{kaupp2010human}.

\section{Formal Problem Statement}

The motivating problem is formally cast as a POMDP, specified by the 7-tuple $\{\mathcal{S,A,T,R},\Omega,\mathcal{O},\gamma\}$.
Let the continuous states of the mobile robot and target in some search environment be $s_{r}$ and $s_{t}$, respectively.  
The joint state space is $[s_{r},s_{t}]^{T} = s \in \mathcal{S} = \mathbb{R}^{N}$, where $N$ is the dimensionality of the combined state space. For the problem considered in this work, each agent is modeled in the 2D plane, with the height of the mobile robot handled via an unmodeled secondary controller. This results in a combined state space with $N=4$. The human sensor has full knowledge of $s_{r}$ at all times as well the belief $b(s_{t}) = p(s_{t}|o_{h}^{1:k},o_{r}^{1:k},a^{1:k-1})$, which is the updated Bayesian posterior pdf given all observations made by the robot $o_{h}^{1:k}$ and the human $o_{h}^{1:k}$ through time $k$. The action space $\mathcal{A}$ is a combination of movement action made by the robot which affect its state $\mathcal{A}_{m}$ and query actions which pull information from the human $\mathcal{A}_{q}$. 

The human has access to a limited set of camera perspectives that show a live feed of the simulated environment, shown in Figure \ref{fig:bothAnnotated}, which they use to gather information about the target's location. They represent this information through sketches $\mathcal{H}$ and an associated spatial information cue (Near, Inside, North, South, etc.). The sketches are semantically labeled spatial areas corresponding to salient features of the space. These labels, along with problem relevant relational indicators, form the semantic codebook from which the query actions $A_{q}$ are drawn. The sketches themselves are defined geometrically, as a set of vertices forming a convex polytope in $\realspace^{2}$. Each sketch also corresponds to a softmax observation model like those used in \cite{burks2019optimal,burks2021collaborative}, which is derived based on the polytopes vertices \cite{sweet2016structured} and is the observation model $\Omega = p(o|s,a_{q})$ for a given query action $a_{q}$. An example softmax model is shown in Figure \ref{fig:reducedFunction}.

The observation set $\mathcal{O}$ is split into robotic detection observations $\mathcal{O}_{r} = \{No  Detection, Detection, Capture\}$ and human answers to query actions $\mathcal{O}_{h} = \{Yes, No, Null\}$, accounting for the possibility of the human not answering. 
Each sketch creates an enlarged query action space $\mathcal{A}_{q}'$ such that $\mathcal{A}_{q} \subset \mathcal{A}_{q}'$, where each new action is a possible query to the human regarding the relative location of $s_{t}$ with respect to the new sketch. For a given sketch labeled $(l)$, with a set of relevant relational indicators, $\mathcal{A}_{q}' = \mathcal{A}_{q} \cup [Relation \times l]$. Thus the robot is capable of asking more questions after a sketch due to the expansion of possible semantic queries.

Finally, the reward function $\mathcal{R}$ is defined as a piece-wise function to incentivize target capture and slightly penalize human queries such that for some distance threshold $\tau$,
\begin{align}
\mathcal{R} = 
\begin{cases} 
      \mathcal{R}_{c} & dist(s_{t},s_{r}) \leq \tau \\
      \mathcal{R}_{n} & dist(s_{t},s_{r}) > \tau, a_{q} = Null \\
      \mathcal{R}_{q} & dist(s_{t},s_{r}) > \tau, a_{q} \neq Null \\
\end{cases}
\end{align}
In this work, the capture reward $\mathcal{R}_{c}=100$, the cost of querying $\mathcal{R}_{q}=-1$, and null query is allowed at no cost for $\mathcal{R}_{n} = 0$.
A POMDP policy $\pi$ for this problem is one which maximizes sum of expected future discounted reward of actions $E[\sum_{k=0}^{\infty}\gamma^{k}\mathcal{R}^k]$, where $\mathcal{R}^k$ is the reward at time $k$.

\section{Technical Approach}

Many existing approaches for augmenting autonomous robotic reasoning with human-robot semantic dialog only allow robots to reason about limited kinds of uncertainties, such as assuming prior knowledge of task environments and conditions that do not change in unforeseen ways. This limits the flexibility and utility of semantic communication for adapting to new or unknown situations.

As shown in Fig. \ref{fig:marquee}, this work features three key technical innovations. Firstly, humans are modeled as ad-hoc sensors that push \emph{multi-modal} semantic data to robots via structured language, enabling robots to update task models and beliefs with information outside their nominal sensing range. These multi-modal inputs from the operator can update the POMDP's own observation models ($\mathcal{O}$) \TROEdits{or action space ($\mathcal{A}$), in addition to} semantic information pertaining to previously abstracted or ignored state information, such as \TROEdits{behaviors which change depending on the target mode}. Such dynamic modeling was expressly forbidden by the techniques used in \cite{burks2019optimal,burks2021collaborative}, which required a static, known POMDP 7-tuple, but are now accessible via ad hoc semantic sensor models.

Secondly, humans can use real-time sketch interfaces to update semantic language dictionaries grounded in uncertain environments, thus dynamically extending the range and flexibility of their observations. Finally, robots actively query humans for specific semantic data at multiple task model modes to improve online performance, while also planning \TROEdits{with foresight} to act with imperfect human sensors. These features effectively enable online `reprogramming' of uncertain POMDPs together with human-robot sensor fusion to support online re-planning in complex environments. These innovations are discussed in turn in each of the following subsections.

\subsection{Leveraging Operators as Sensors}

\subsubsection{Human Querying and Data Fusion}
%
%
In this work it is assumed that the human can act as either a passive semantic sensor, which volunteers information without request whenever possible, or as an active semantic sensor, which can be queried by the robot to provide information on request. This allows the human to proactively offer information they intuit to be useful, and for the robot to directly gather what it perceives to be the most vital information. However, as with many other sensor modalities, the use of a human sensor forces the consideration of sensor accuracy and responsiveness. An accurate characterization of these two attributes is necessary to determine the influence of the human's information on the robot's belief of the target location. In addition, the required value of such information needs to bear the cost of a query action when the sensor may not respond. 

The human's responsiveness to queries is based on a static a priori known \TROEdits{availability} value $\xi$, such that, for all human observations $o_{h}$ at all times, $p(o_{h} \neq None|s) = \xi$. 
This leads to an additional observation $o_h=None \in \Omega_{h}$. 
It is also assumed that the human has a static a priori known accuracy rate $\eta \in [0,1]$. \TROEdits{Within the space of all possible human observations $o_h \in \mathcal{O}$, a single observation $i$, such as `North', can be mislabeled as a different observation $j$, such as `West'. We therefore account for the probability of an operator being wrong by redistributing the probability to the inaccurate observations,} 
\begin{align}
    &\bar{p}(o_{h} = j|s) = p(o_{h} = j|s) \cdot \eta \\
    &\bar{p}(o_{h} = (i \neq j)|s) = \\  &\frac{1}{|\Omega|-1} p(o_{h} = j|s) \cdot \eta \cdot p(o_{h} = (i \neq j)|s) \nonumber
\end{align} 
The notation $\bar{p}(o|s)$ and $\bar{p}(s'|s,a)$ denote models used during online execution, in contrast to the nominal distributions $p(o|s)$ and $p(s'|s,a)$. 
This parameterization of accuracy 
ignores similarity between any two observations. For example, the probability mass associated with an observation `East' being inaccurate is not allocated primarily to `Northeast' and `Southeast', but instead evenly allocated amongst all other possible observations. While this simplifies implementation and allows comparative testing of accuracy \TROEdits{values}, other models may yield more realistic results; e.g. 
linear softmax model parameters \cite{burks2018closed,burks2019collaborative,Ahmed-SPL-2018},
can be used to determine the likelihood of mistaking semantic labels given $s$. 

This work focuses on human operators represented as active sensors. This requires an explicit dependency on actions to be included in the observation model, as well as additional actions which trigger this dependency. 
Just as POMDP observations $o \in \Omega$ can be typically modeled as $o \sim p(o|s)$, 
human responses to robotic queries $a \in \mathcal{A}_{q}$ can be modeled as $o_{h} \sim p(o_{h}|s,a_{q}), \  a_{q} \in \mathcal{A}_{q}, \ o_{h} \in \Omega_{h}$. 
These additional query actions can be introduced into $\mathcal{A}$ in one of two ways. Casting them as exclusive options, where either movement or a query can be pursued but not both, minimally expands the action space to the \TROEdits{union of queries and movements, $\mathcal{A} = \mathcal{A}_{m} \cup \mathcal{A}_{q}$}. But this can be limiting in situations which benefit from rapid information gathering about models and states together. 
Instead, queries are cast as inclusive actions, in which every time step permits any combination of movement and query, $\mathcal{A} = \mathcal{A}_{m} \times \mathcal{A}_{q}$. This can drastically increase the size of the action space, but allows rapid information gathering. A similar technique was successfully applied by \cite{burks2019collaborative} for the more limited static semantic \TROEdits{codebook} case. 

The robot's on-board sensor produces a single categorical observation per time step $o_{r}$, which must be fused with $o_h$. We assume the robot's sensor is independent of the human observations given the state such that,
\begin{align}
    p(s|o_{r},o_{h}) = \frac{p(s)p(o_{h},o_{r}|s)}{p(o_{r},o_{h})}  
    \propto p(s)p(o_{r}|s)p(o_{h}|s)
    \label{eq:stateUp1}
\end{align}

\begin{figure}[t]
\centering	
    \includegraphics[width=.25\textwidth]{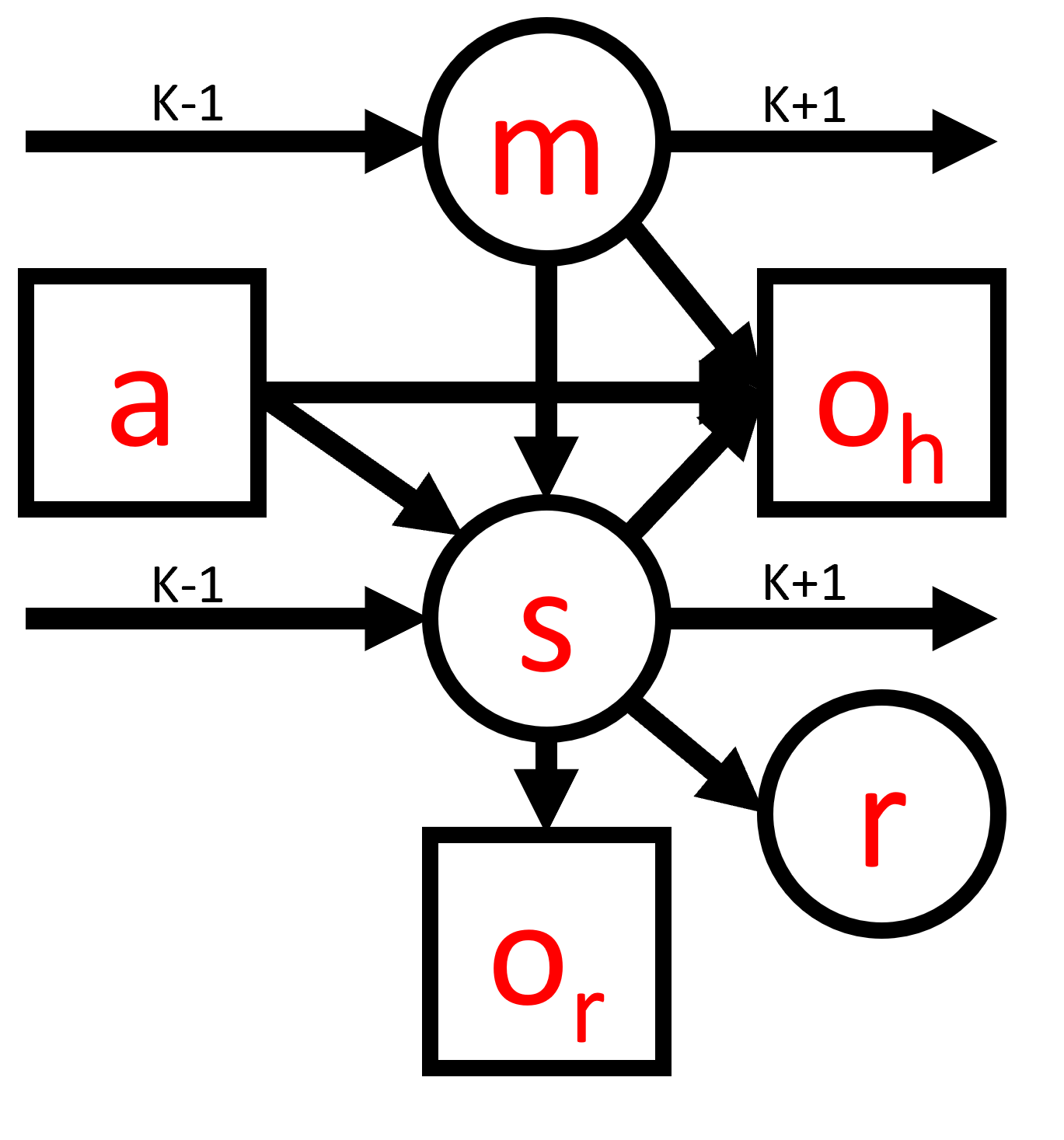}
	\caption{Graphical model for the proposed POMDP}
    \label{fig:graphicalModels}
    \vspace{-0.25in}
\end{figure}

Fig. \ref{fig:graphicalModels} summarizes the probabilistic dependencies for the POMDPs describing the motivating search problem in Fig. \ref{fig:bothAnnotated}. While all observations are state dependent, some now depend on actions chosen by the policy. Also, query actions have no effect on the state, and are pure information gathering actions. In these new POMDPs, a combined action might be ``Move to North, and Ask human if target South of Lake", which may return a combined observation of ``Target is Far from me, and human says target is not South of Lake". So in this case $a=\{North, South/Lake\}$, and $o=\{Far,No\}$.

\subsubsection{Multi-Modal Active Information Gathering}
As described in the motivating problem of Fig. \ref{fig:bothAnnotated}, the target can transition between two high level modes, either constrained to the roadway or unconstrained on foot. These two modes carry their own transition dynamics, and separately influence other state's transitions. For example, the target cannot transition into the road-constrained state without first being near a roadway, but can move much more quickly between spatial states once they have. To accommodate stochastic switching dynamics, it is generally convenient to include a finite number of discrete mode states $m$ as shown in Fig. \ref{fig:graphicalModels}, which
dictate alterations to the target state transition models, 
\begin{align}
    \bar{s} = [s^{T},m] \label{eq:stateAdd},  
    m' \sim p(m'|m) \ 
    s' \sim p(s'|s,a,m'). 
\end{align}
This additional discrete state variable can be easily handled by POMCP and other Monte Carlo tree search approximations, which rely on generative black box simulators and support edits to $\mathcal{T}$ and $\mathcal{O}$.    
Offline approximations based on
switching-mode POMDPs  \cite{brunskill2010planning} can also accommodate hybrid dynamics, but require stationary models. 
Such hybrid model extensions also open the door to active semantic queries and specific human observations $o_h$ pertaining to $m$ (e.g. `Is the target still traveling on the road'), which can greatly enhance the Bayesian belief for $m$. 



With this modification to the generative model, the probability of different state transition models being in effect can be explicitly represented via particle approximation as
\begin{align}
    P(m=x) = \frac{1}{N} \sum_{n=0}^{N} \textbf{1}(m_{n} = x),
\end{align}
where $\textbf{1}(m_{n}=x)$ is an indicator function applied to the mode of particle ($n$).

For human responses modeled as $p(o_{h}|m)$, the belief update equation becomes 
\begin{align}
    &p(s|o_{r},o_{h}) = \sum_{m} p(s,m|o_{r},o_{h})  \label{eq:modeUp}\\
    &= \sum_{m} p(s|o_{r},o_{h},m)p(m|o_{r},o_{h}) 
    = \sum_{m} \frac{p(o_r|s)p(m|s)p(o_h|m)}{p(o_h)p(o_r)} \nonumber
\end{align}
If the mode $m$ is treated as an discrete addendum to the continuous state vector $s_{t}$, as in Equation \ref{eq:stateAdd}, this equation reduces to Equation \ref{eq:stateUp1}.
The bootstrap particle filter used in Algorithm 2 approximates this belief update through the use of weighted particles. 



\subsection{Using Structured Language and Sketches for Intuitive Interaction}
\subsubsection{Language-based and Sketch-based Semantic Soft Data}
While addressing scenarios of unknown/dynamic environments the human primarily acts as an auxiliary and imperfect semantic information source that can communicate with the robot at any time via either one of two interfaces, described in Fig. \ref{fig:marquee}. 
The first interface allows the human to compose linguistic statements that are parsed and interpreted as target state observations, via the positive and negative semantic informational statements described in Section IIA. 

The second interface addresses the novel problem of incorporating the human information into the robot's map knowledge. Since new map information cannot be obtained from the robot's sensor to augment the codebook, 
the human may instead do this by providing labeled 2D free-form sketches, which each depict a spatially constrained region on a 2D map display (as in Fig. \ref{fig:marquee} with `Pond' and `Trees').
While similar 2D sketch interfaces have also been developed for robotics applications \cite{skubic2003sketch,boniardi2015robot,shah2012sketch,ahmed2015fully}, 
their use in the context of multi-modal data fusion for planning under uncertainty represents a novel contribution. 

Building on \cite{burks2019collaborative}, the codebook is automatically augmented with the labels of new landmarks/references, and the corresponding spatial sketch data can also be used to generate suitable soft data likelihood models for the linguistic statement interface (see Fig. \ref{fig:marquee}). However, unlike in \cite{burks2019collaborative}, human sketches here may also provide direct information about probabilistic state transition models, to constrain how the robot and target may traverse certain map areas. 

In this work, the model for the human sensor is assumed to be known in advance. That is, given a sketch $h_{i}$, the autonomy is capable of interpreting human responses to queries regarding  $h_{i}$ according to a probabilistic likelihood function. Regarding the scope of human soft data in this work, such assumed models, including those for accuracy and availability, are further assumed to be static, and are not learned or adapted online. Furthermore, while the online planning approach in the following sections could in principle be applied to changing reward functions without modification, here only changes to the observation function is considered. Such changes are taken `as is' by the autonomy, rather than maintaining a belief over the uncertain model parameters as in previous work \cite{ross2008bayes}. 

Finally, the example problems in this work assume the semantic codebook to be empty at the start of the problem, and built up from scratch by the human using semantically labeled sketches. However, the techniques for online planning and sketch processing apply identically in problems for which a prior codebook is available yet incomplete. In any case, building the codebook from scratch implies a slightly more difficult problem due to the potentially lacking information available at the start of the problem, and thus provides a more compelling case for human sketch input.

\subsubsection{Sketch Generation and Integration}
A sketch, or drawing in the 2D plane, begins as a set of points $\{P\}$, whether with a pen, touchscreen, mouse, or other modality. The set $\{P\}$ is tagged by the human with a natural language label $(l)$, and any applicable meta information $\delta$. The communication of the label and meta information in this work is accomplished using fillable text boxes and radio button selection of a pre-selected set of meta data options, but could also be acquired entirely from natural language through word association methods such as Word2Vec \cite{mikolov2013efficient} or other deep learning methods.

From the set $\{P\}$, the ordered \TROEdits{vertices} of a convex hull $v_{i} \in \{V\}$, where $\{V\} \subset \{P\}$ is obtained using the Quickhull algorithm. $\{V\}$ is progressively reduced using Algorithm \ref{alg:HullReduce} until it reaches a predefinied size by repeatedly removing the point contributing the least deflection angle to the line between its neighbors, calculated via the Law of Cosines for the vertex pair vectors $\overrightarrow{v_{i-1}v_{i}}$,
\begin{align}
    \Theta(v_{i}) = \arccos{\left[ \frac{\overrightarrow{v_{i-1}v_{i}} \cdot \overrightarrow{v_{i}v_{i+1}}}{\|\overrightarrow{v_{i-1}v_{i}}\| \|\overrightarrow{v_{i}v_{i+1}}\|} \right]}
\end{align}
in a procedure inspired by the Ramer-Douglas-Peucker algorithm \cite{douglas1973algorithms}.

{\scriptsize
\begin{algorithm}[h!]
\caption{Sequential Convex Hull Reduction}
\begin{algorithmic} \label{alg:HullReduce}
\STATE \textbf{Function:} $REDUCE$
\STATE \textbf{Input:} Convex Hull $\{V\}$, Target Number $N$

\IF{$\{V\} == N$}
    \STATE return hull
\ENDIF
\FOR{$v_{i} \in \{V\}$}
    \STATE $\Theta(v_{i}) \leftarrow angle(v_{i-1},v_{i},v_{i+1})$
\ENDFOR
\STATE $\{V\} \leftarrow \{V\} \setminus argmin_{v} \Theta(v)$
\STATE return REDUCE(hull,N)
\end{algorithmic} 
\end{algorithm}
}
This heuristic was chosen as a proxy for maximizing the area maintained by the reduced hull, but in practice other heuristics could also be used. The reduced set is then used as the basis for softmax model synthesis \cite{sweet2016structured}, in which $M$ points define a final sketch $h_{i}$ associated with a softmax function with $|h_{i}| = M+1$ classes. In general, these are the single interior class contained within the sketched polygon region and $M$ exterior classes distributed outside the sketched polygon region, each associated with a relational indicator using the methodology described in this work's Supplementary Material. 

A limitation of sequential hull reduction is the need to pre-define a set number of points at which to stop the progressive reduction. Ideally, the ``natural" number of points needed to maximize some criteria could be chosen on a sketch by sketch basis. 

\begin{figure}[t]
    \centering
    \begin{subfigure}{0.43\textwidth}
        \includegraphics[width=0.9\textwidth]{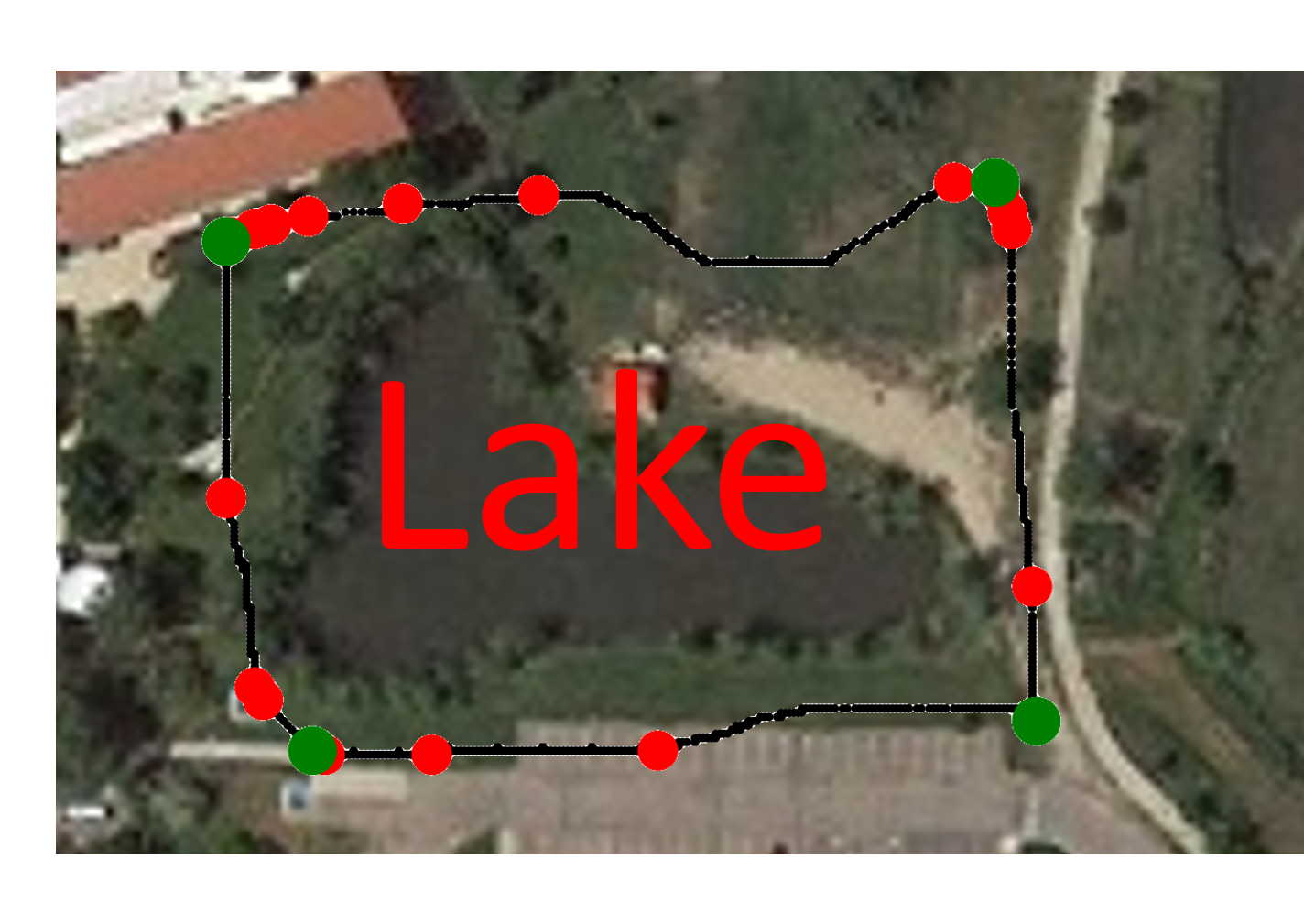}
    \end{subfigure}
    ~
    \begin{subfigure}{0.43\textwidth}
        \includegraphics[width=.9\textwidth]{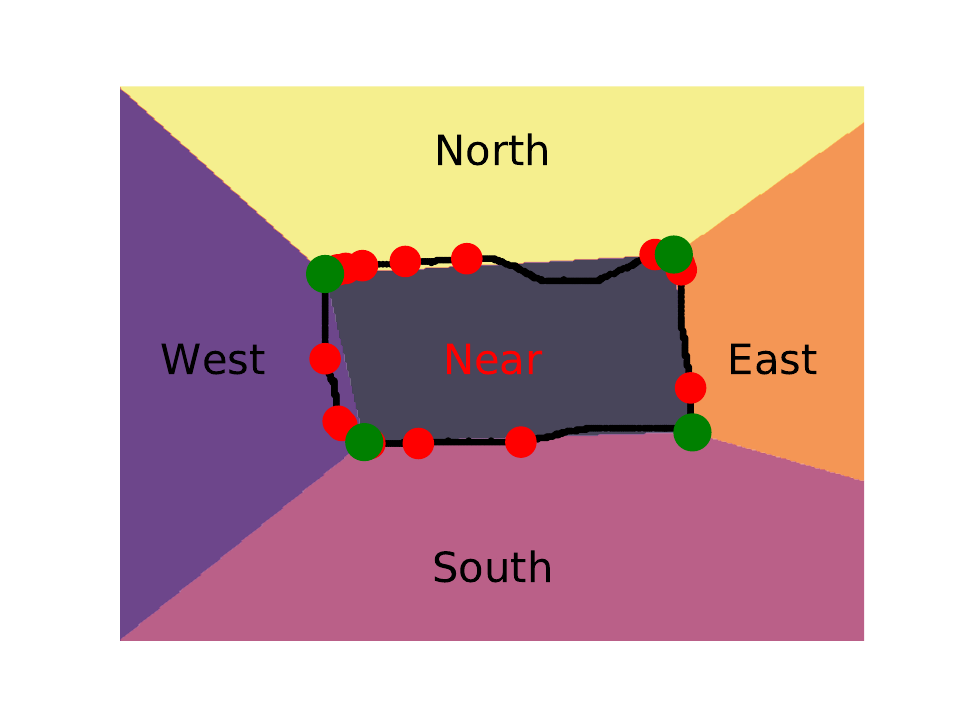}
    \end{subfigure}
    ~
    \begin{subfigure}{0.35\textwidth}
        \includegraphics[width=.9\textwidth]{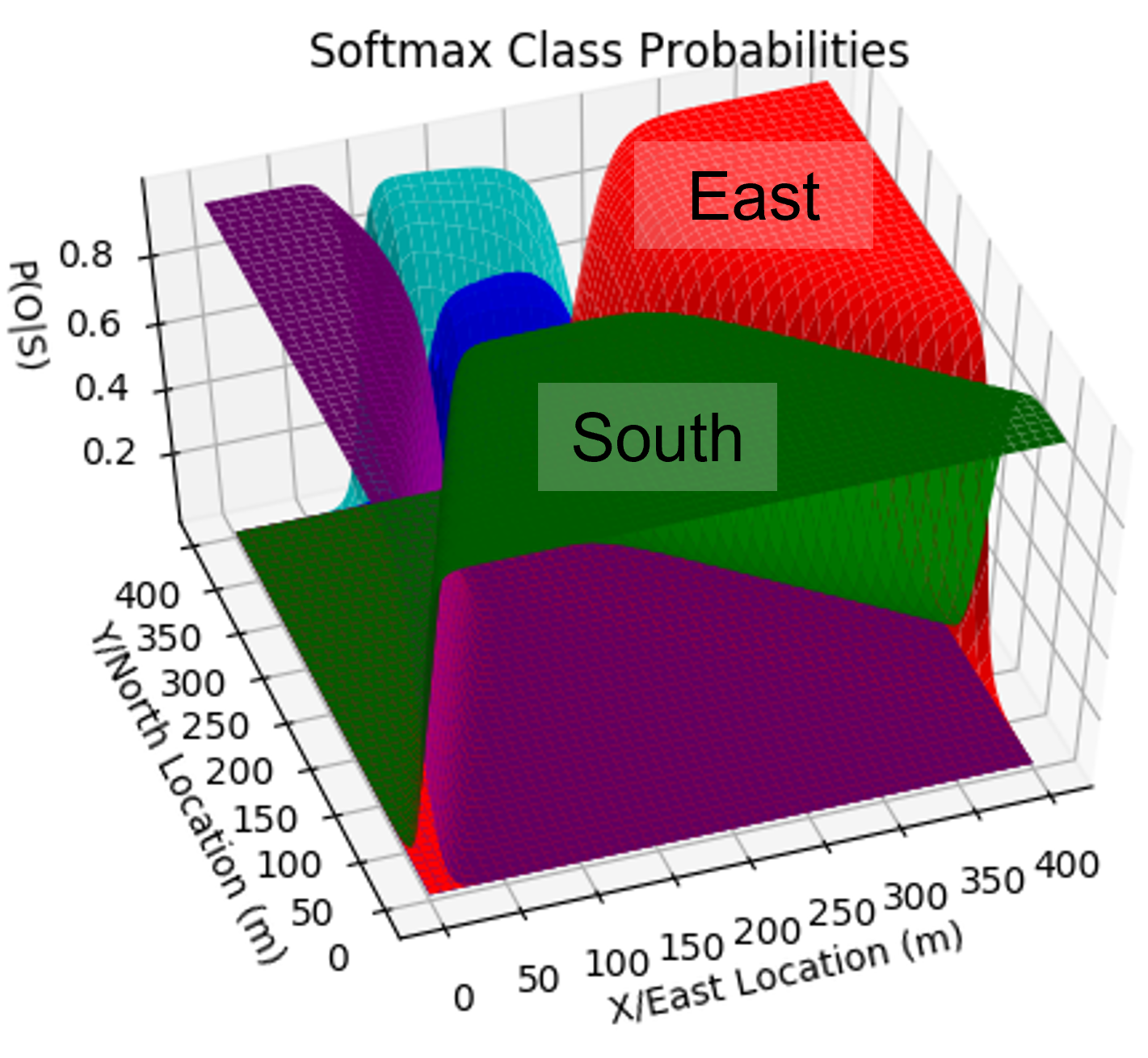}
    \end{subfigure}
    \caption{Top: Convex hull verticies (red), reduced to 4 points with sequential hull reduction (green). Middle: Softmax function and labels resulting from reduced points. Bottom: A example of Softmax likelihoods mapped across the state space.}
    \label{fig:reducedFunction}
    \vspace{-0.1 cm}
\end{figure}

An example sketch is shown in Fig. \ref{fig:reducedFunction}, where the initial input consists of 661 points, shown in black, making a roughly rectangular shape. The Quickhull algorithm is applied to find 21 points, shown in red, defining a convex hull on the set of points. These points are then used as input to Algorithm \ref{alg:HullReduce}, which further reduces the number of points to the 4 points shown in green. From these 4 points, a softmax model consisting of 5 classes, a ``Near" class and 4 relational indicators (`North', `South', `East', `West'), is then synthesized. \TROEdits{Where range observations are also in use alongside semantic bearing, an additional interior class "Inside" can be constructed as a subset of the "Near" spatial region.} Semantic observations can now be constructed using the label $(l)$ given by the human when they made the sketch in the form, $o = \text{``The Target is $relation$ of $label$."}$ The area inside the sketch $h_{i}$ is now modelled as subject to any modifying meta data $\delta_{i}$ provided by the human. This introduces a scaling factor, in which:
\begin{align}
    E[|s'-s|] = \delta_{i}, \ \forall s \in h_{i}
\end{align}

The query action set $A_{q}$ and observation function $\Omega$ are modified, introducing the newly minted sketch $h_{i} \in \mathcal{H}$ and its corresponding softmax function such the one shown in Figure \ref{fig:reducedFunction}. The observation set $\mathcal{O}$ remains the same in this work, but through the modified $\Omega = p(o \in \mathcal{O}|s,a)$ observations corresponding to the new query actions in $A_{q}$ are now available. The availability $\xi$ and accuracy $\eta$ parameters of the human collaborator are static across any model revisions, and their effect is folded into changes to $\Omega$. In this way, both parameters are accounted for in planning through their effect on the observation \TROEdits{likelihood}. 

In principle the human could provide nothing more than the sketched points $\{P\}$, with no label. However, the robot will still need to indicate specific sketches to the human in queries, potentially requiring it to resort to referring to ``Sketch 1" vs ``Sketch 2". Semantic labeling of sketches ultimately leads to better understanding on the human's part of the robot's queries and, though not implemented here, enables the use of the semantic label to automatically infer additional metadata using natural language processing techniques. \TROEdits{This metadata could be used to modify the robot's transition function, such as pointing out a difficult sand trap.}

\subsection{Dynamic POMDP Modeling}

\TROEdits{Having updated the POMDP with a revised observation likelihood through a human sketch, a POMDP planner can produce a new optimal policy. In this case, both the true and internal model of observations change with the sketch, as the human is able to specify their model directly through the sketch. The changed model implies that the robot must solve a different but related POMDP after each human sketch. With this in mind, we next consider viable approximations for solving such POMDPs.}


{
\begin{algorithm}[b!]
\caption{Planning with Human Model Updates}
\begin{algorithmic}[1] \label{alg:UpdateFlow}
\small 
\STATE $B_{k} =$ Particle\_Set(Size = N)
\REPEAT
\STATE $[a_{m},a_{q}]$ = POMCP($B$) (Ref. \cite{silver2010monte})
\STATE $s \sim p_{k}(s'|s,a_{m})$ (unknown state)
\STATE $o_{r} \sim \bar{p}_{k}(o_{r}|s)$ (robot sensor observation)
\STATE $o_{h} \sim \bar{p}_{k}(o_{h}|s,a_{q})$ (human query answer)
\STATE $B_{k+1} =$ Bootstrap\_Filter($B_{k},a_{m},a_{q},o_{r},o_{h}$)
\IF{New Human Sketch (h)}
    \STATE l = $h.l$, (human assigned label)
    \STATE $\Omega_{k+1} = \Omega_{k} \cup (l \times [Near,E,W,N,S])$ 
\ENDIF
\UNTIL{Scenario End}




\end{algorithmic}    
\end{algorithm}
}

The Partially Observable Monte Carlo Planning (POMCP) algorithm proposed in \cite{silver2010monte} is particularly promising due to its use of generative `forward' models and online any-time characteristics. 
While the original implementation of POMCP uses an unweighted particle filter for belief updates, the authors of \cite{egorov2017pomdps} note that, for problems with even moderately large $\mathcal{A}$ or $\mathcal{O}$, a bootstrap particle filter allows for more consistent belief updates without domain specific particle reinvigoration.
This weighted particle filter approach coincidentally also provides a solution to the dynamic modeling problem. Each belief update is carried out using the most up-to-date model, while changes to the model only affect future timesteps. 
As model alterations require solving a different POMDP after each sketch, this approach allows each planning phase to be treated as the start of a brand new POMDP solution. Our procedure for carrying out POMCP planning while handling dynamic model updates from a human is detailed in Algorithm \ref{alg:UpdateFlow}. 

\begin{figure*}[h!]
\centering	
    \includegraphics[width=.7\textwidth]{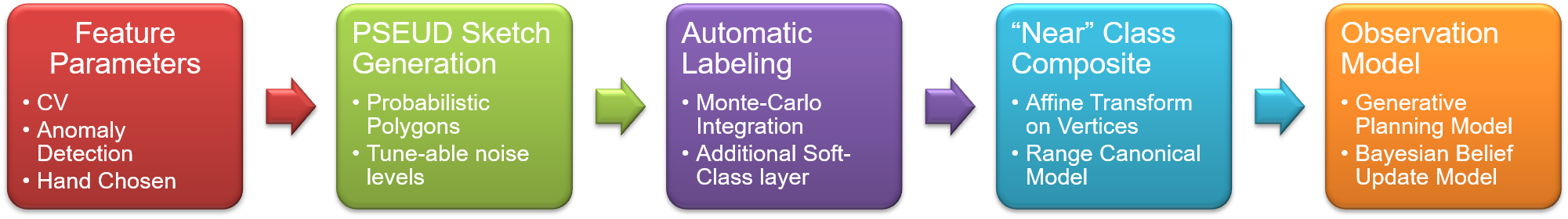}
	\caption{Synthetic sketch generation and preparation pipeline.}
    \label{fig:sketch_pipeline}
\end{figure*}

\begin{figure}[h]
\centering	
    \includegraphics[width=.4\textwidth]{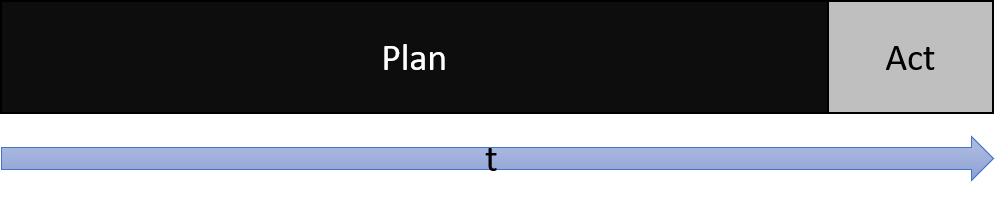}
	\caption{A typical offline POMDP time allocation for planning and acting.}
    \label{fig:OfflinePlan}
\end{figure}

Implementation of an online POMDP solver such as POMCP in a time-limited system raises an additional complication, regardless of human involvement. A typical offline POMDP solver computes a policy prior to execution and requires significantly less time to execute actions recommended thereby, as shown in Figure \ref{fig:OfflinePlan}.

\begin{figure}[h]
\centering	
    \includegraphics[width=.4\textwidth]{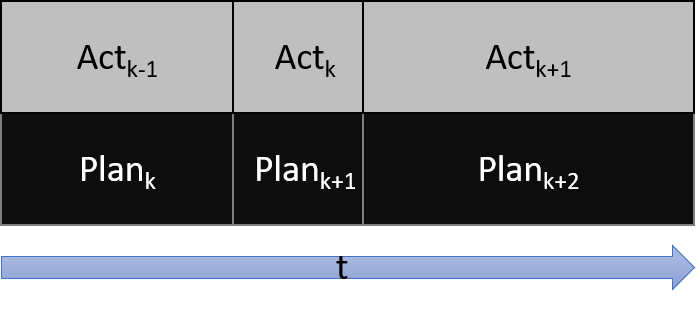}
	\caption{Our framework allocates planning and acting over irregular time durations.}
    \label{fig:IrrPlan}
\end{figure}

Online planners like POMCP interweave planning and execution during runtime, leading to an alternating time allocation structure. This structure, when implemented directly on a robotic platform, leaves the robot immobile while planning and non-cognizant while acting. Planning for timestep $k+1$ cannot take place until the observation at timestep $k$ has been fused into the belief, as POMCP operates on the current belief as the root node of its planning tree. If this challenge could be overcome, it would be advantageous to operate according to a structure where the planning for timestep $k+1$ takes place during the execution of action $k$. Additionally, given that POMCP is limited in this work to a maximum decision time, it would be ideal to allocate exactly as much decision time as action $k$ takes to execute, as shown in Figure \ref{fig:IrrPlan}. An approach to achieving this structure is proposed here, known as Predictive Tree Planning. 



If such planning can take place only after receiving an observation, the tree structure of POMCP can be leveraged to pre-plan by predicting which observation will be received. For a given action $a_{k} \in A_{k}$ and belief $b_{k}$, the observation model $p(o|s)$ is used to direct tree samples to nodes with history $b_{k}a_{k}o$. The weighted proportion of samples in nodes $[b_{k}a_{k}o_{1},b_{k}a_{k}o_{2},...,b_{k}a_{k}o_{|\Omega|}]$ represents the distribution $p(o_{k}|b_{k},a_{k})$,
\begin{align}
    p(o_{k} = o | b_{k},a_{k}) = \sum_{s_{k} \in b_{k}} w_{s} p(o_{k}=1|s_{k},a_{k})
\end{align}
which is the observation probability distribution for the current belief and action. In order to spend $\Delta t(a_{k})$ time planning during the execution of $a_{k}$, $|\Omega|$ plans can be pursued, each allocated $\Delta t(a_{k}) p(o_{k} = o | b_{k},a_{k})$ time. That is to say, after carrying out a belief update assuming a future observation, we generate a POMCP policy using time in proportion to the probability of receiving said observation. Upon receiving the true observation at the end of the action execution phase, the autonomous system can immediately begin executing the action from the policy corresponding to that observation. That is, having generated $N_{p}$ policies $[\pi(o_{k+1}=0),\pi(o_{k+1}=1),...,\pi(o_{k+1}=N)$, and received the observation $o_{k+1} = 1$, the policy $\pi(o_{k+1} = 1)$ can be used to find action $a_{k+1}$ without delay.

It must be noted that although this approach works well for all of the example application problems considered here (as there are only two observations the drone can receive during a movement action) it faces significant scalability issues. Creating and choosing between significantly larger numbers of predictive planning trees reduces the accuracy of the policy approximation sharply, especially if many observations hold similar likelihoods. This can be addressed in future work through a likelihood threshold to filter out rare observations, combined with a heuristic to determine actions when such rare observations are seen. Alternatively, observations could be collected into meta-observations when they lead to similar beliefs, echoing the discretization of observations spaces seen in \cite{porta2006point}.

\section{Simulated HARPS Results}

\subsection{Observation Model Generation Pipeline}

\begin{figure}[h!]
\centering	
    \includegraphics[width=.48\textwidth]{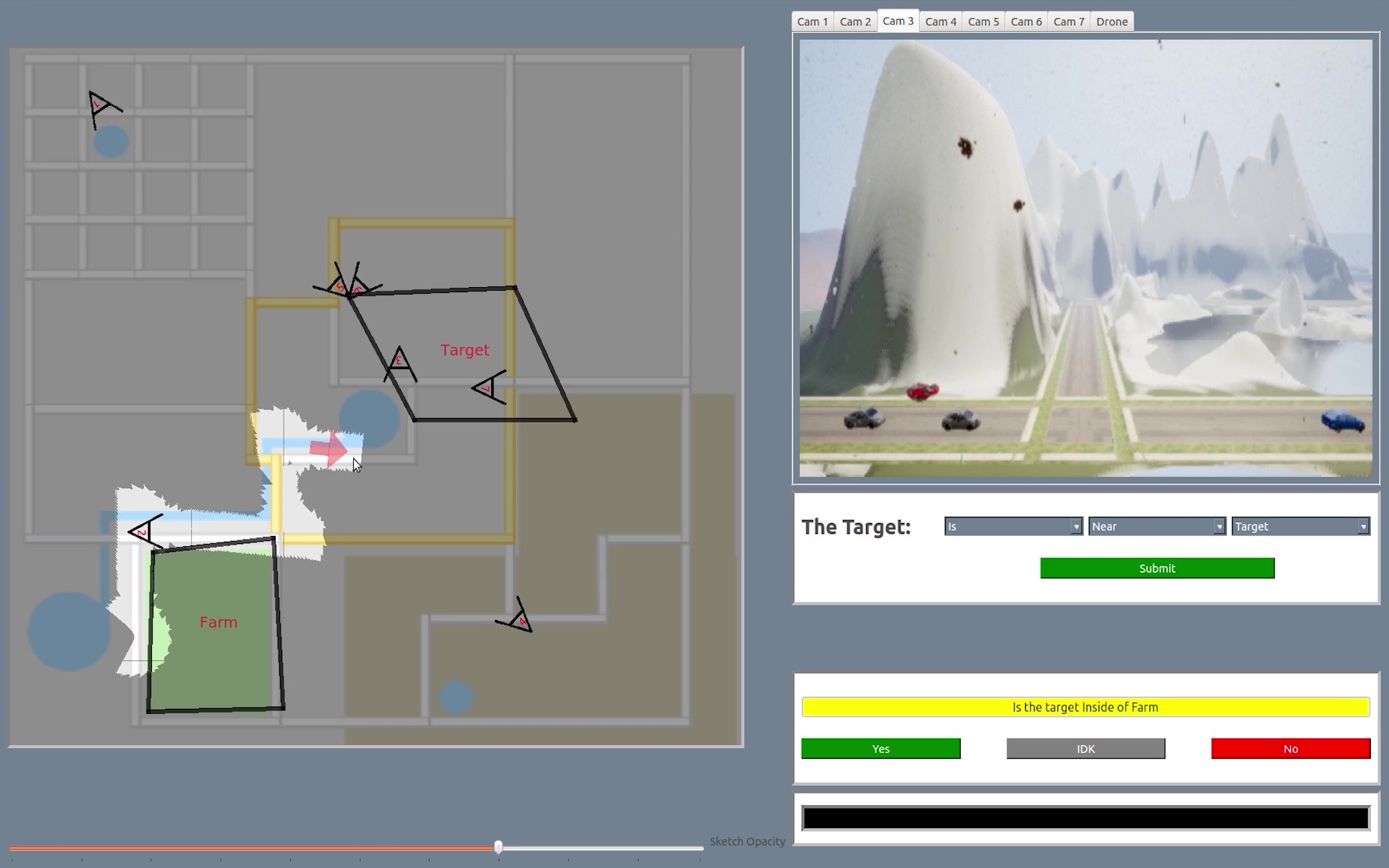}
	\caption{HARPS user interface shows an overhead view of the operational area on the left, and live camera views from limited locations on the right. The structured language interface is available on the bottom right.}
    \label{fig:HARPS_interface}
    
\end{figure}

The HARPS simulation environment was developed in Unreal Engine 4 \cite{unrealengine} to provide physics-based testing of POMDP tracking problems with human input. A drone, controlled through ROS \cite{ros} using the Microsoft Airsim API \cite{airsim2017fsr} attempts to localize, track, and intercept a ground based target within an outdoor environment, while interacting with a human supervisor through the sketch based interface shown in Figure \ref{fig:HARPS_interface}. This PYQT5 based interface implements the information querying functionality with response buttons and drop down lists, similar to the interface in \cite{burks2021collaborative}, but populates the semantic codebook within from user sketches drawn directly on the map. The human can view the space either through a selection of security cameras, as shown in Figure \ref{fig:bothAnnotated}, or through camera on-board the UAS.

To assess the ability of an online POMDP to work together with a broad range of human sensing characteristics to complete this task, a parameter study was first performed on a slightly simplified version of the HARPS environment built in Python 3.5. The parameter study examined a variety of parameterized human sensor characteristics for generating sketches and observations via the methods detailed in this work's Supplementary Material.
This allowed for large batch testing to examine the POMDP's effectiveness and potential weak points prior to live human subject testing (Sec. VI). 

A single end-to-end pipeline was developed to produce data for simulated testing. This pipeline, shown in Figure \ref{fig:sketch_pipeline}, generates synthetic sketches using specified probability distributions on variables of interest such as number of sides, regularity, and size. Then the sketches are processed in a Monte Carlo Auto-labeling routine to assign semantic labels before a range model is extracted. The final result, an observation likelihood specified by a set of softmax models, can then be used in both the belief update and policy search processes. This allows testing to cover the full range of possible human inputs in simulated experiments with minimal human input beyond specified ranges of sketch input parameters, and allows minimal effort construction of likelihood models in human testing using only the input sketch. Details on individual elements of the pipeline are given in this work's Supplementary Material.

\subsection{Experimental Setup}

\TROEdits{We simulate the interaction over 250 trial runs, during which the drone was given $t_{max} = 600s$ maximum flight time, simulating the on-station search time of low-cost commercially available platforms. Runs reaching this time limit without capturing the target were considered failures. Each trial used randomly generated robot and target start points.} 
Rather than pose actions as simple directional directive (``move North" and ``move South") as in previous experiments \cite{burks2021collaborative,burks2019optimal}, HARPS uses a graph-based POMDP navigation approach, which designates intersections, or nodes, of the local road network as destinations corresponding to movement actions in $\mathcal{A}_m$ from neighboring nodes. Specifically, the robot was allowed to choose any neighboring node, or neighbors of neighbors, from its previous action:
\begin{align}
    A_{m}^{t} = \cup n_{t+1} \in Neighbors(Neighbors(n^{t})) 
\end{align}
This opens the door to variable time actions, and the use of Predictive Tree Planning techniques discussed in Sec. IVc. 
The robot is allowed to move at an average speed of $15 \frac{m}{s}$, with a standard deviation of $1m$ for all actions, while the target follows the switching Markov dynamics model in the road network problem, using an average speed of $20 \frac{m}{s}$ and standard deviation of $5m$ while on the road and $5 \frac{m}{s}$ and standard deviation of $1m$ while off-road. 
Query actions $A_{q}$ were generated in real-time using the PSEUD algorithm, detailed in this work's Supplementary Material. 

HARPS uses a conical detection model for the aerial robot with a 30 degree viewcone $v_{c}$ similar to that used in \cite{burks2021collaborative}, with observation set:

\begin{equation}
    \mathcal{O} = \{None, Detected, Captured\}
\end{equation}

and observation likelihood model with a 98\% accuracy:

\begin{align}
    \Omega(Captured) = p(o=Captured|s) = 0.98, \\ \forall s | dist(s_{t},s_{r}) \leq \tau \ \& \ s \in v_{c} \nonumber \\ \nonumber \\
    \Omega(Detected) = p(o=Detected|s) = 0.98, \\ \forall s | \tau \leq dist(s_{t},s_{r}) \leq 2\tau \ \& \ s \in v_{c}\nonumber \\ \nonumber \\
    \Omega(None) = p(o=None|s) = 0.98, \\ \forall s | 2\tau \leq dist(s_{t},s_{r}) \ \& \ s \in v_{c} \nonumber
\end{align}

As HARPS is applied to a relatively larger scale environment, with state space $\mathcal{S}$ comprising a 1000m x 1000m area, the capture distance threshold $\tau$ is set to 75m. However, the restriction of detection and capture to the area under observation by viewcone $v_{c}$ limits the ability of the robot to detect targets it is not directly facing as a result of movement actions $A_{m}$.

The reward function was given a large positive reward $\mathcal{R}_{c}$ for the capture of the target within $\tau = 75m$, and a small penalty of $\mathcal{R}_{q}$ for asking the human a question.   

A variety of variables were tested to examine their influence on the robot's ability to capture the target. These included the parameters accuracy, availability, and sketch rate, which dictate the human's ability to provide information, as well as the use of the Predictive Tree Search method which modified the planner's treatment of future observations.  
All statistical testing in the simplified HARPS environment was conducted using binomial significance tests on the ratio of successful captures by a given method when compared to a control.

\subsection{Simulated Human-assisted vs. Unassisted Performance}

\begin{table}[t]
    \centering
    \begin{tabular}{||c|c|}
    \hline
        Input Type & Capture Rate \\
        \hline
        Human & 0.82 \\
        No Human & 0.63 \\
    \hline
    \end{tabular}
    \caption{The effect of human input in the Simulated HARPS Environment over 250 trials was significant with $p<0.001$.}
    \label{tab:simHARPS_Human}
\end{table}

Table \ref{tab:simHARPS_Human} compares the ability of the drone to intercept the target with and without human assistance. Each case was tested with N=250 independent trials, with significance being determined via Binomial Test. It is particularly important to note that the robot is able to locate and intercept the target in the majority of cases using the POMDP planner even without simulated human intervention. With or without the human's information, the online POMDP is still approximating an optimal policy for the information it has available. Far from minimizing the improvement shown in the ``Human" case, the significant $(p<0.05)$ discrepancy with the ``Nonhuman" case emphasizes that even an otherwise effective algorithm can still benefit greatly from the introduction of additional human information. \TROEdits{The human case was run with a default accuracy and availability of 90\%.} 

\begin{figure}[h!]
\centering	
    \includegraphics[width=.48\textwidth]{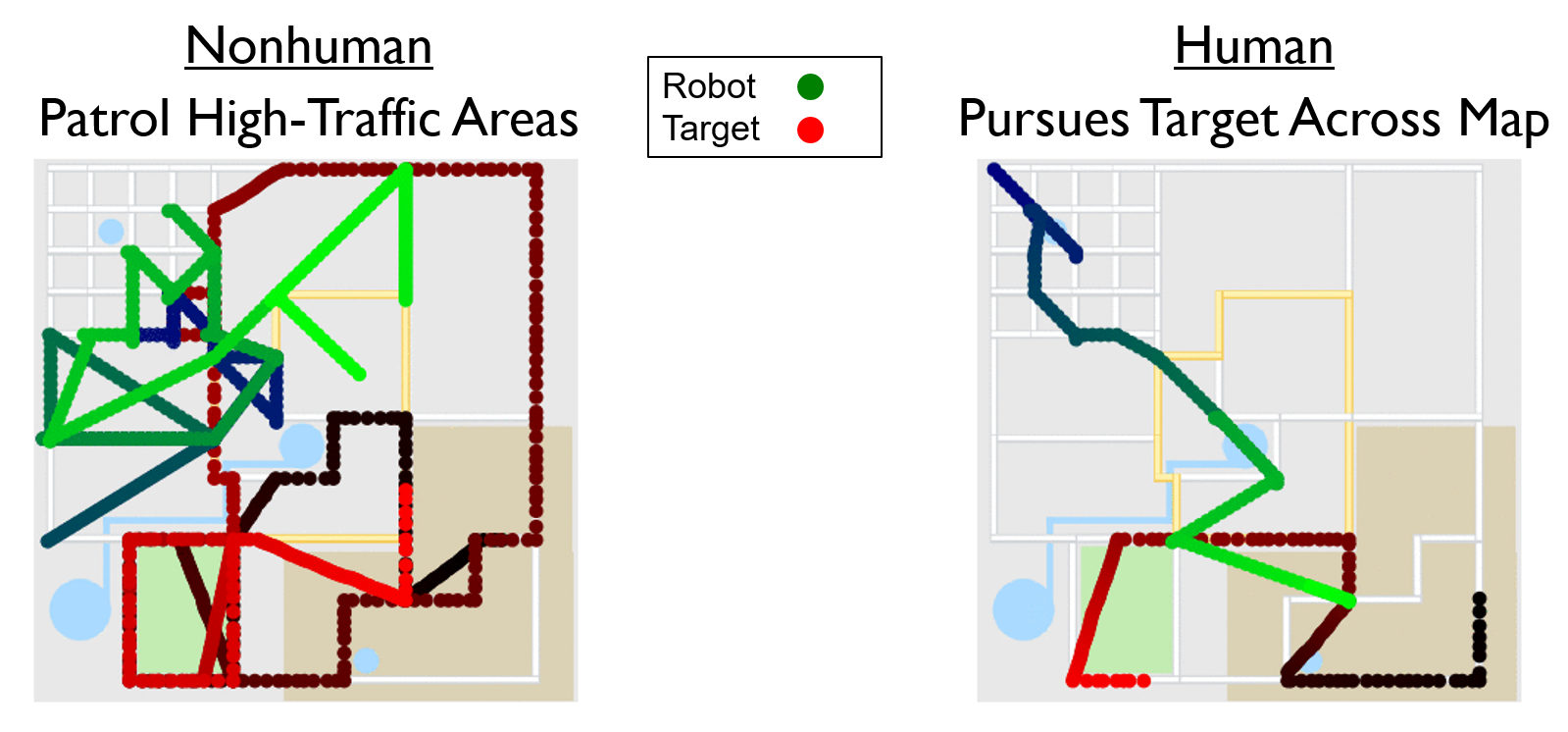}
	\caption{Representative example traces of the robot (green) and target (red) showcasing patrol behavior of non-human HARPS case on the left, and pursuit behavior with human information on the right.}
    \label{fig:vignettes}
\end{figure}

Qualitatively, the introduction of human information resulted in similar behavioral changes as in \cite{burks2021collaborative}. For example, in Figure \ref{fig:vignettes}, without human input, either sketches or semantic observations, the robot engages in a broad ranging patrol behavior, prioritizing well connected nodes in the graph and eventually capturing the target as it ventured through a high traffic area. However, the human case of Figure \ref{fig:vignettes} shows a more directed pursuit behavior, wherein the robot moves quickly towards the target area indicated by the human before firmly localizing and capturing the target. Note, this example does not imply all human cases were so direct, nor all non-human cases so wandering. Rather, it represents the general behavior of each approach. As will be seen in Section VI with real human data, the general range and effectiveness of patrol/pursuit behaviors are correlated with the types of state belief updates produced by different human sensors. 

\subsection{Predictive Tree Search vs Blind Dynamics Updates}
\begin{table}[t]
    \centering
    \begin{tabular}{||c|c|}
    \hline
        Planning Type & Capture Rate \\
        \hline
        Blind & 0.76 \\
        Predictive Tree Search & 0.85 \\
    \hline
    \end{tabular}
    \caption{The effect of Predictive Tree Search in the simulated HARPS Environment over 100 trials was significant with $p=0.03$.}
    \label{tab:simHARPS_Predict}
\end{table}

\TROEdits{In Table \ref{tab:simHARPS_Predict}, the Predictive Tree Search method is tested against a baseline ``blind" real-time planner.} Each case was tested with $N=100$ independent trails, with significance being determined via Binomial Test. In the blind case, planning is also carried out during the execution of the previous action, but using a belief predicated only on the expected dynamics update. This is opposed to the predictive method which probabilistically allocates planning time among multiple possible observations to produce predicted measurement updates in addition to the dynamics update. Of note, the blindness of the baseline method only applies to a single step, as the full measurement update still occurs after the end of the current action, so it is at most one step behind. Here the Predictive Tree Search approach produces significantly $(p<0.05)$ more captures by aircraft operating \TROEdits{within the} 10 minute search restriction. This may be attributable to the fact that the UAS moves at a slower speed than the target on the road, and therefore timely arrival of a single positive measurement can have an out-sized effect on the ability of the drone to capture the target before it escapes the local area.

\subsection{Human Accuracy}
\begin{figure}[h!]
\centering	
    \includegraphics[width=.45\textwidth]{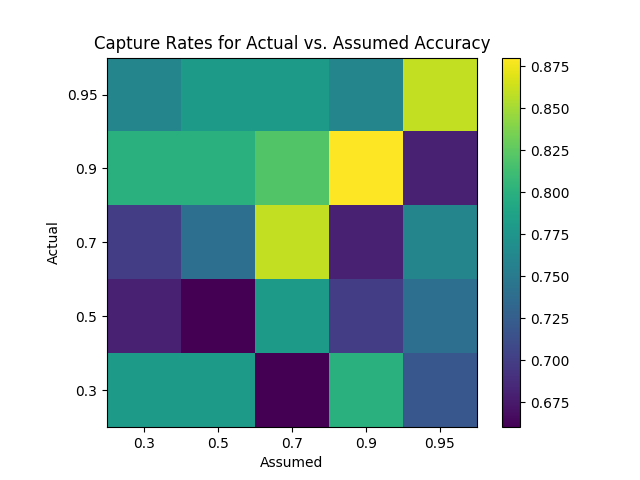}
	\caption{Effect of True Human Accuracy vs Modeled Accuracy in Simulated HARPS Environment.}
    \label{fig:simHARPS_AccData}
    
\end{figure}

After establishing the general effect of a simulated human collaborator on the effectiveness of the POMDP, specific attributes likely to vary between live subjects were examined. First, human accuracy was tested, using values in the set: $[0.3, 0.5, 0.7, 0.9, 0.95]$. That is, for a value of $0.7$, the human would respond to questions correctly 70\% of the time, and incorrectly 30\% of the time. These same values were tested as applied to the robot's model of the human, allowing the difference between the true and assumed sensor dynamics to be examined. The full results of these tests, carried out for $N=50$ independent trails for each combination of assumed and actual accuracy, are show in Figure \ref{fig:simHARPS_AccData}. 
A human operating at 50\% accuracy acts as a random observation generator, and therefore does not produce significantly more captures than the earlier ``Nonhuman" case. The 30\% case allows the human observations to be taken in negative form, still allowing useful information to be passed along.  

\TROEdits{The effects of both assumed and actual accuracy can also been seen in the rows and columns of Figure \ref{fig:simHARPS_AccData}.} The average true accuracy shows similar results to the match case, with the averaging over assumptions compressing the range of results, while the assumed accuracy displays no clearly significant effect. Naturally there can be limits to such data in marginalized form, as examining the combined effects indicates a slight advantage to a more pessimistic view of the human's accuracy. That is, the upper-left triangle of results in which the human is more accurate than assumed performs slightly better on average than the lower-right. 



\subsection{Human Availability}
\begin{figure}[h!]
\centering	
    \includegraphics[width=.45\textwidth]{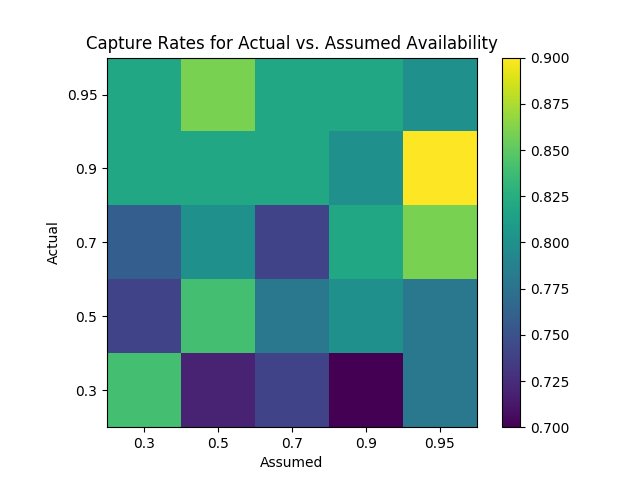}
	\caption{Effect of True Human Availability vs Modeled Availability in Simulated HARPS Environment}
    \label{fig:simHARPS_AvailData}
    
\end{figure}

Similar to accuracy, human availability was also tested across the same range of assumed and true values. Given the same size of $N=50$, the results for the matched true to assumed availability did not indicate a strong effect. However, examining the marginal result distributions in Figure \ref{fig:simHARPS_AvailData}, there is a clear positive trend line for true human \TROEdits{availability}, while the effect of the robot's assumption of the human's accuracy is minimal.




\subsection{Sketch Rate}
\begin{figure}[h!]
	\centering	
		\includegraphics[width=.45\textwidth]{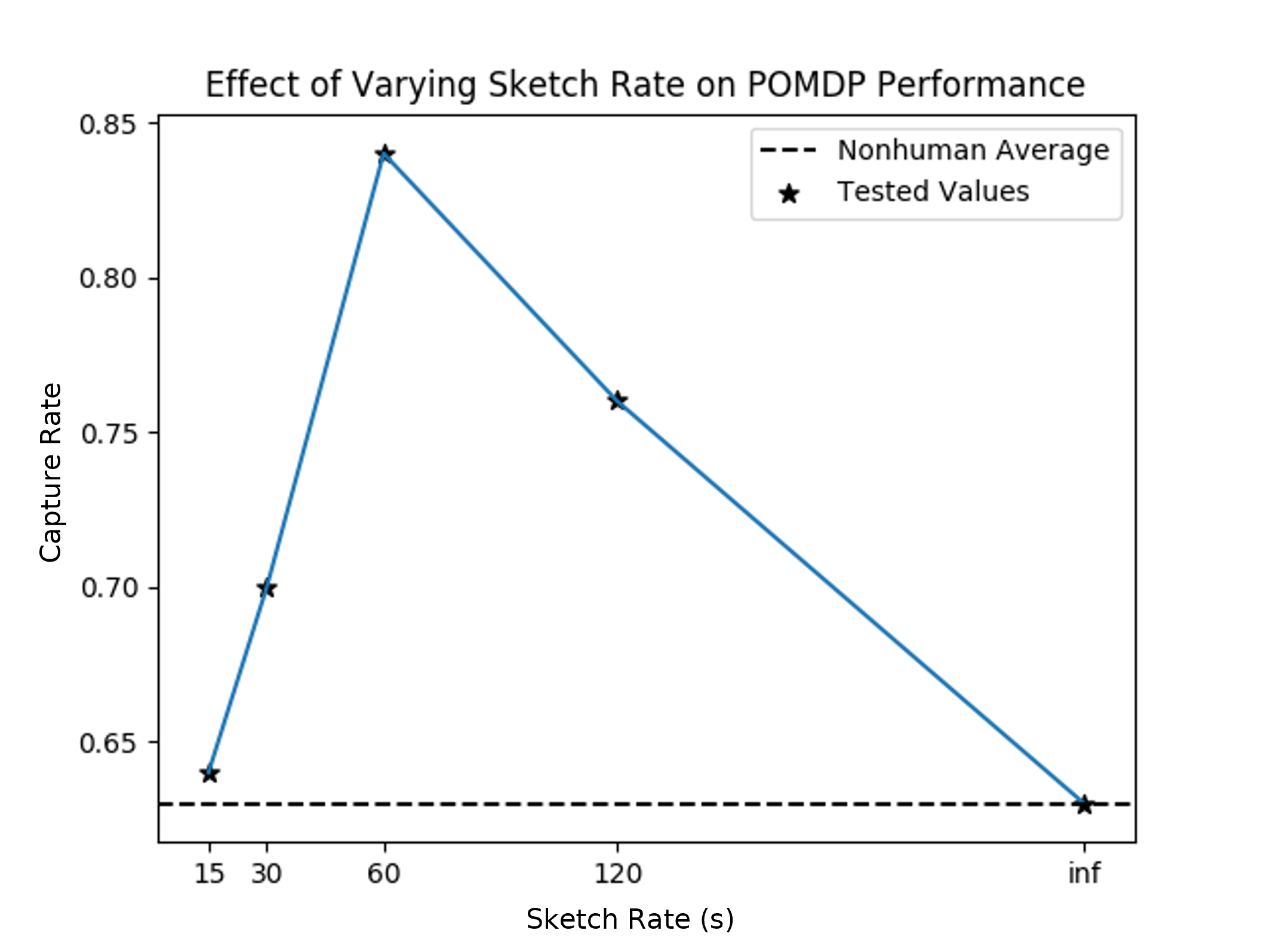}
	\caption{Effect of Sketch Rate in Simulated HARPS Environment}
    \label{fig:simHARPS_SketchRate}
\vspace{-0.05 in}
\end{figure}

Finally, the rate at which a human inserts additional sketch entries into the robot's semantic \TROEdits{codebook} was examined. Simulated humans drew a sketch at a rate of $[15, 30, 60, 120]$ seconds. For completeness, a trial was run in which the human was present but drew no sketches and introduced no new semantic information. As expected, this case of an infinitely long average sketching rate performed identically to the Unassisted baseline. As shown in Figure \ref{fig:simHARPS_SketchRate}, there exists an optimal rate at which sketches provide a sufficiently diverse semantic \TROEdits{codebook} without over complicating the POMDP with additional actions to consider. This optimum, which peaked around one sketch per minute in this scenario, is likely highly problem dependent and represents a major consideration for any implementation of this work with live human collaborators. As additional sketches primarily modify the planning problem by increasing the number of actions to consider, this optimal rate represents the total volume of new environmental information that can be optimally reasoned over. It is possible that, having established for a particular problem the near-optimal sketch rate or volume, human collaborators could be coached or trained to adhere to it. 

\subsection{Summary of Baseline Simulation Results}

These results taken together showcase the ability of the online POMDP planner to adapt to the human it is given, rather than requiring a collaborator with specific training or knowledge. Certainly more accurate and available humans are of more use, but in no case tested across a range of simulated humans did the robot do significantly worse than the non-human baseline. This indicates that in the worst of cases the POMDP is able to mitigate the effects of a poor human, while maintaining its ability to effectively collaborate with others.

\section{Human Subject Studies}
An IRB-approved user study \footnote{University of Colorado Boulder IRB Protocol \# 20-0587, approved 29 Jan 2021} was run to test the effectiveness of the proposed approach using the same environment presented in Section III. The goal of the study was to evaluate the performance of the human-robot team with different communication abilities and compare the difference to an approach where the robot searches without human assistance. Communication modes include a \textit{passive} operator that opportunistically provides observations to the robot, an \textit{active} operator that can be directly queried by the robot, and a mixed, \textit{both} approach where the operator provides information and responds to queries.  

The study was devised as a within-subjects experiment, which tested a total of 36 subjects with varying levels of experience with autonomy and UAS. Subjects were tasked with assisting the drone in capturing the target in each of three scenarios. They could observe the landscape for the target, or signs of the target, using seven different camera views distributed across the environment as seen in Figure \ref{fig:bothAnnotated}. Sketches could be added on an accompanying 2D map at any time, which also displayed the drone’s location and previous path. 

In the \textit{active} only scenario, the subject could volunteer information only when requested by the robot. For example, the robot could ask \say{Is the target North of the Farm?}, to which the user could respond \say{Yes}, \say{No} or \say{I don’t know}. The passive only scenario allows the subject to provide information without request whenever possible. This modality offered the subject the ability to provide positive or negative information about the target (is/is not), relative location (Near, Inside, North, South, East, West) and the associated sketch (\say{The target is inside the mountains}). Finally, the active/passive scenario combined the capabilities in the previous two other scenarios, allowing bi-directional communication between the robot and subject where the robot queried the user and the subject provided additional information as they saw fit.

Analysis of the results from the simulated user studies leads us to posit the six following hypotheses for the human subject studies. \textbf{H1-H4} relate to objective measures whereas \textbf{H5} and \textbf{H6} correspond to subjective measures:

\textbf{H1}: Respectively incremental human input for the three scenarios, from active only, passive only, and active/passive will improve task performance.

\textbf{H2}: There exists a favorable frequency of human sketch input, such that too many sketches result in inefficient algorithmic planning and performance and too few sketches do not provide sufficient localization of the target. 

\textbf{H3}: A more effective user, defined as having high levels of accuracy $\eta$ and availability $\xi$, will result in improved task performance.

\textbf{H4}: Subjects with better situational awareness, defined as those who switch to more camera views over time, will be more accurate.

\textbf{H5}: Subjects will have incrementally higher respective workloads for active only, passive only, and active/passive scenarios.

\textbf{H6}: Subjects will report higher levels of trust towards the robot with more interaction, with more accurate subjects reporting higher levels of trust.

\textbf{H1} investigates the performance of the algorithm in context of the unique modes available in an interaction. \textbf{H2}, \textbf{H3} and \textbf{H4} aim to evaluate the algorithm's ability to handle a range of user types. \textbf{H5} and \textbf{H6} will provide insight into the user's subjective view of the interaction with the robot across the respective modes.

\subsection{Experimental Procedure}
Upon recruitment to the study, subjects were given a briefing regarding the purpose and scope of the experiment. The experimenter walked subjects through the interface and showed a brief video of the interaction. Once a subject read through the IRB approved consent form, they were assigned a subject ID number and given the opportunity to experiment with the interface before starting the three trials. For each subject, the scenario ordering and target location was randomized \TROEdits{between three initial target and robot start points} to account for learning improvements. For each scenario, subjects were given a maximum of 15 minutes (900 seconds) to find the target with the drone. \TROEdits{The maximum time was increased from 10 minutes after preliminary human trails, to allow for data collection.} After each scenario, the subject filled out a questionnaire to rate their subjective workload and trust observations. Once all three scenarios had been finished, the subject was paid and dismissed.

\subsubsection{Objective Measures}
The objective measures used in this study can be broken down into metrics for performance, defined by time to capture (TTC), and user characteristics, which includes their accuracy, availability, sketch rate, and view rate. \TROEdits{Unless specifically broken out by scenario mode (active, passive, or both), the results for each subject are combined.}

\underline{Performance:}
    \begin{itemize}
        \item Time to Capture: Defined as the time taken from the start of the simulation until the time where the robot came within 75m to the target. If the drone failed to capture the target, then the TTC is not reported, and the scenario updates the capture rate for the respective run. 
        \item Capture Rate: The number of successful captures divided by the total number of runs for each category of scenario $(n = 36)$.
    \end{itemize}

\underline{User Characteristics:}
    \begin{itemize}
        \item Accuracy: For each observation provided by the subject, either passively or actively provided, an accuracy value, representing the truthfulness of the statement, can be reported using a softmax approximation as defined in Section IV. Equation  \ref{eq:softaccuracy} shows how the total accuracy is calculated where $p_o$ is the softmax derived probability of the target being in the assigned label and $p_{max}$ is the maximum possible probability for that specific observation and target location. For comparison, a heuristic compass approximation is also calculated and whose total value is calculated according to equation \ref{eq:compassaccuracy}. In this equation, $p_o$ similarly refers to the probability of the target being in the assigned label, although the definition constrains this value to $0.33 \le x \le 1$.
        \begin{align}
            \mathbf{accuracy_{softmax}} &= \frac{1}{N} \sum_{o\in O} \frac{p_o}{p_{max}} 
            \label{eq:softaccuracy}\\
            \mathbf{accuracy_{compass}} &= \frac{1}{N} \sum_{o\in O} 
            \begin{cases} 
                        1 &  p_o \neq 0 \\
                        0 & p_o = 0\\
            \end{cases}
            \label{eq:compassaccuracy}
        \end{align}
        \item Availability: The total questions answered divided by the total number of questions asked by the robot.
        \item Sketch Rate: The number of sketches provided by the user divided by the scenario time. If the target was captured, then the TTC is used as the scenario time, otherwise the duration of the scenario (15 minutes) is used.
        \item View Rate: The rate at which a subject looks at different camera views, is given by the total number of camera views divided by the scenario time.
    \end{itemize}

\subsubsection{Subjective Measures}
At the conclusion of each trial, subjects were given questionnaires to assess their subjective workload, measured by the NASA RTLX survey, and their trust in the robot, using the Multi-Dimensional Measure of Trust (MDMT) survey. 

The NASA Task Load Index (NASA-TLX) has been successfully applied across a range of domains to gauge user workloads \cite{TLXmain}. We used a modified version known as the Raw TLX (RTLX) to simplify the questionnaires for the user. The RTLX is the most common modification to the classic TLX survey \cite{TLXraw}. It dispenses with the relative weightings found in the classic TLX survey and simply takes the average of six questions that were reported on a scale from 0 (very low) to 7 (very high).

The MDMT survey was developed as a novel measurement tool to asses human-robot trust \cite{MDMT}. It gathers perceived trust across four differentiable dimensions, Reliable, Capable, Ethical, and Sincere, which are respectively gauged by a series of four questions. Each dimension can be further categorized into the broader factors of Capacity Trust and Moral Trust, of which we only focused on Capacity Trust. Each question is evaluated on an 8-point scale from 0 (Not at All) to 7 (Very) and can be ignored by the user if they deem it irrelevant. The total reported trust for each trial is calculated as the average of the answered questions.

\subsection{Results}
Participants were recruited from across the University of Colorado Boulder student body, as well as members of the local community. All 36 subjects were between the ages of 18-65 and reported 20/20 or correctible to 20/20 vision. While two subjects were professional UAS operators involved in surveying, and three subjects were active in search and rescue capacities, others came from a background with no UAS, aviation, or operational experience. There was also a range of experience with decision making algorithms as 15 (42\%) of subjects reported no subject matter background and 9 (25\%) reported high levels of familiarity. As a baseline metric of performance, the simulation was run without an active user for $n=40$ trials.

\subsubsection{Performance}

\TROEdits{\textbf{H1} predicted that modes with more human input would improve overall task performance, the results of which are shown in Table \ref{tab:Performance}}. The capture rate for each of the respective study modes shows a clear increase increase in capture rate with increasing levels of human interaction. When no human is present, the robot captured the target only 47.5\% of the time, compared to the active human (77.8\%), passive human (88.9\%), and active/passive human (97.2\%). Using a binomial significance test, all user cases are shown to be significant to $p<0.001$ when compared to the robot only situation. When the target was captured by the robot, the mean TTC is slightly higher for the robot only case and does not have a significant variation across all of the interaction modes. However, it is important to consider that the mean time to capture does not account for instances where the robot failed to capture the target. While the mean TTC is not statistically significant on its own, the doubling in capture rate from the robot only mode to the active/passive human mode, and incremental improvement across interaction modes, substantiates \textbf{H1}. 


\begin{table}[t]
    \centering
    \begin{tabular}{||c||c|c|c|c|}
    \hline
        Mode & Active & Passive & Both & No Human  \\
        \hline
         TTC (s) & $303 \pm35$ & $313 \pm 40$ & $305 \pm 36$ & $338 \pm 61$ \\
         Capture Rate & 77.8\% & 88.9\% & 97.2\% & 45.5\% \\
         Workload & $3.4\pm0.2$ & $3.3\pm0.1$ & $3.2 \pm 0.2$ & N/A \\
         Trust & $3.6 \pm0.3$ & $3.6\pm0.2$ & $4.0 \pm 0.2$ & N/A \\
    \hline
    \end{tabular}
    \caption{Performance metrics for human subject studies including time to capture (TTC), time to capture standard deviation, and capture rate.}
    \label{tab:Performance}
\end{table}

\begin{figure}
    \centering
         \includegraphics[width=0.45\textwidth]{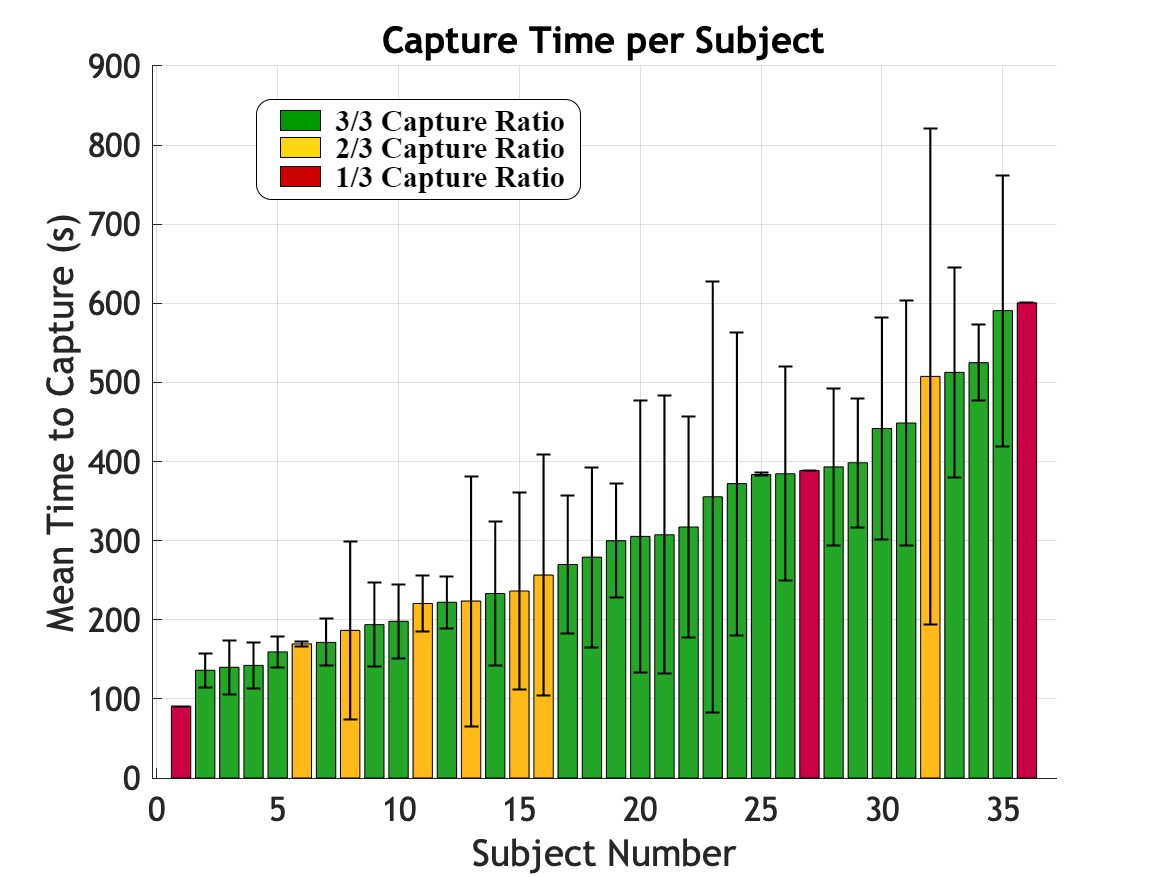}
         \caption{Subject specific mean time to capture with associated standard deviation.}
         \label{fig:Human_SubjectPerf}
         \vspace{-12pt}
\end{figure}

\begin{figure*}
\centering 
         \includegraphics[width=0.9\textwidth]{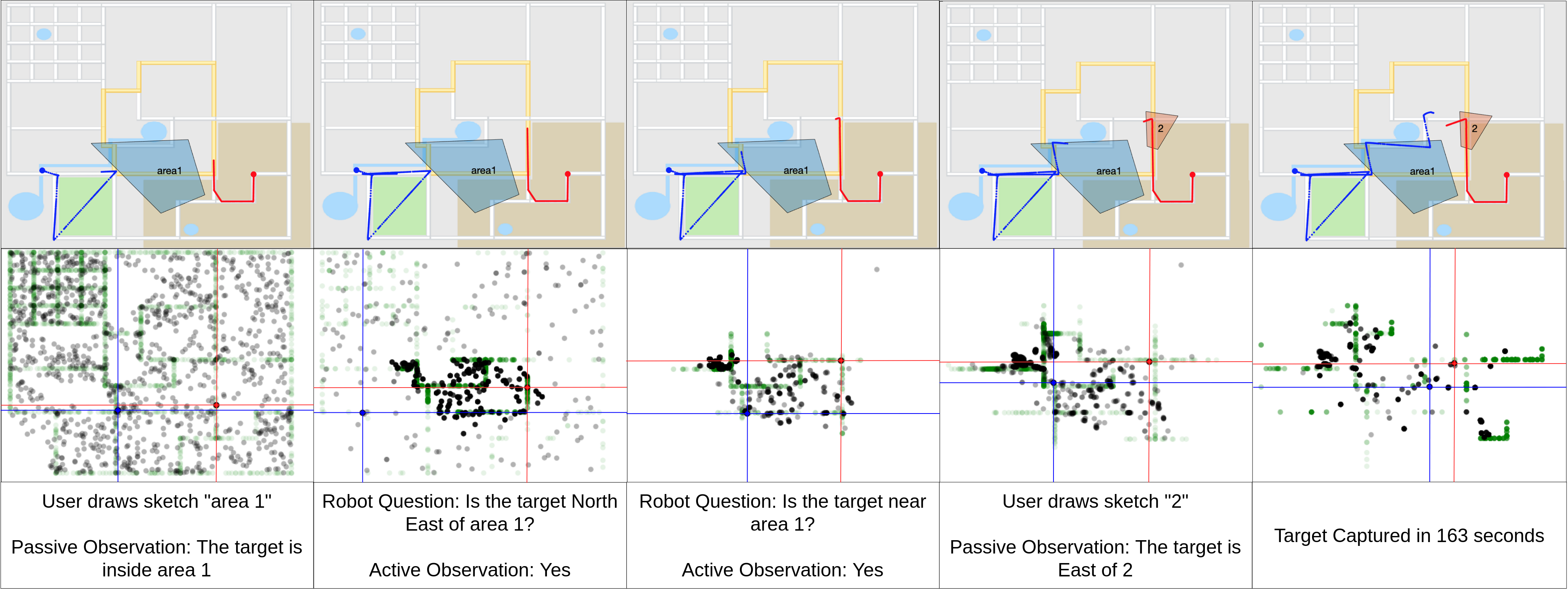}
         \caption{Example of a successful interaction using the active/passive mode.}
         \label{fig:GoodUserBoth}
\end{figure*}

\begin{figure*}
    \centering
         \includegraphics[width=0.75\textwidth]{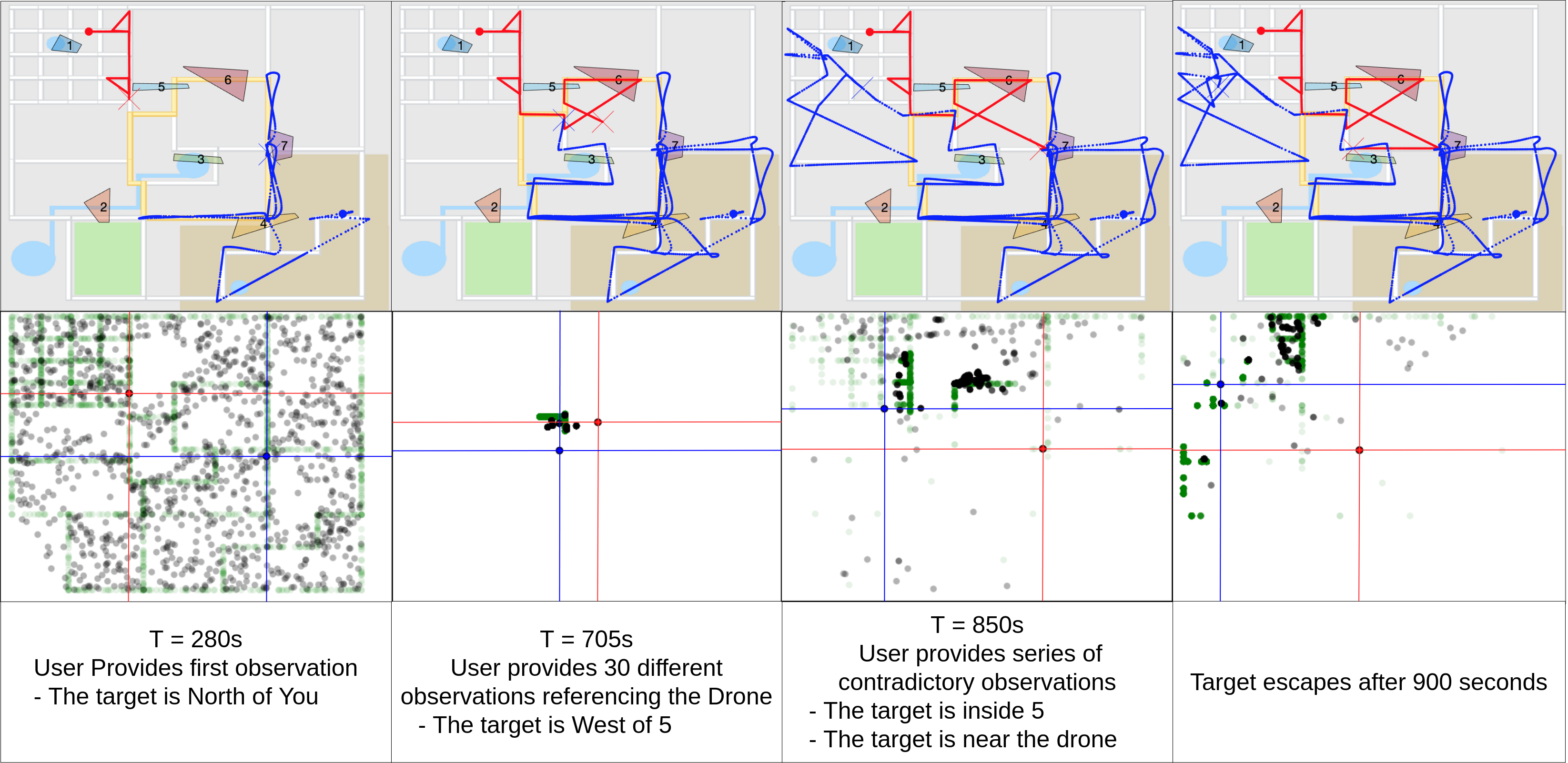}
         \caption{Example of a poor interaction using the passive mode.}
         \label{fig:BadUserPush}
\end{figure*}


\subsubsection{User Type}
The tested subjects compromised a wide range of interaction styles and consistencies. Figure \ref{fig:Human_SubjectPerf} shows each subject's performance across all three scenarios sorted by mean TTC. Some users, such as subjects 2, 3, 4, and 5 showed consistently low mean TTC. Alternatively other users, such as subjects 32, 33, and 35 showed repeated failures to capture the target and a wide standard error across their respective scenarios. There were other instances, such as with subject 1, where the target was captured primarily through luck for one mode of interaction and failed to capture to target on the other two modes. These trends are indicative that the method by which the user interacts with the agent can have a substantial effect in their resulting performance. Analysis of sketch frequency, user accuracy, availability, and \TROEdits{number} of camera views attempt to quantify the particular variables that influence the success of each interaction.


\TROEdits{\textbf{H2} predicted that there would be an optimal frequency of sketches, such as that found in the simulated experiments.} Users drew anywhere from 1 to a maximum of 14 sketches throughout the simulation, some examples of which are shown in Figure \ref{fig:CuriousSketches}. While some users preferred to draw a series of sketches prior to searching for the target, others would only draw sketches when the target had been observed. Unlike the simulated HARPS environment, there was no clear correlating trend for user performance and the number of sketches that had been drawn. \TROEdits{In fact, the scenarios with three of the top four highest number of sketches drawn (14, 13, and 11 sketches) failed to capture the target, with the robot demonstrating poor planning behavior, such as repeated routes. The inefficient planning stems from an exponential growth in action space with each added sketch.} As subjects were able to capture the target in less than two minutes with as few as one sketch and as many as nine sketches, there does not prove to be a favorable number or frequency of sketches that results in optimal performance, therefore we do not have enough evidence in favor of hypothesis \textbf{H2}.

\begin{figure}
     \centering
    \includegraphics[width=0.45\textwidth]{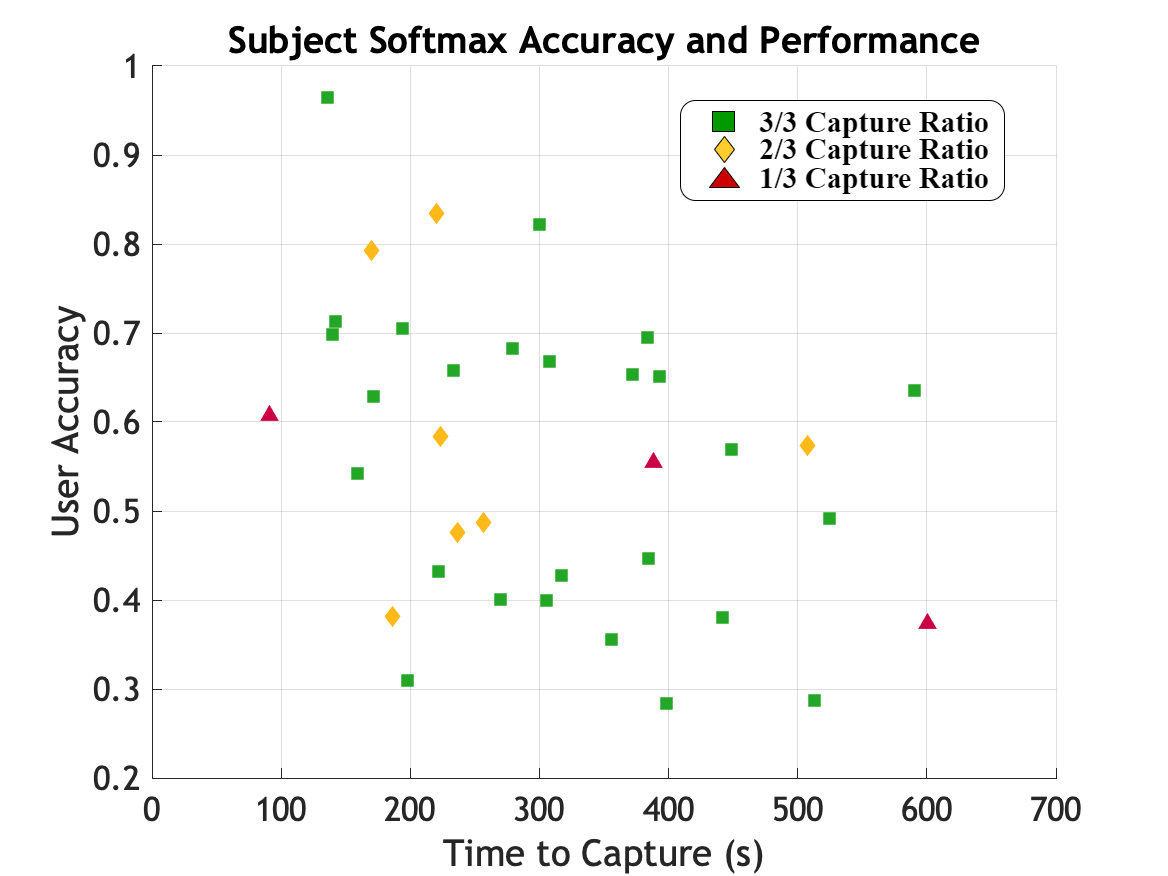}
    \caption{Average user accuracy compared to their mean time to capture. Colors indicate capture rates of $3/3$, $2/3$, and $1/3$ for green squares, yellow diamonds and red triangles respectively. }
    \label{fig:Human_Accuracy}
    \vspace{-17pt}
\end{figure}

\begin{figure}[h]
    \vspace{4pt}
    \centering	
    \includegraphics[width=0.45\textwidth]{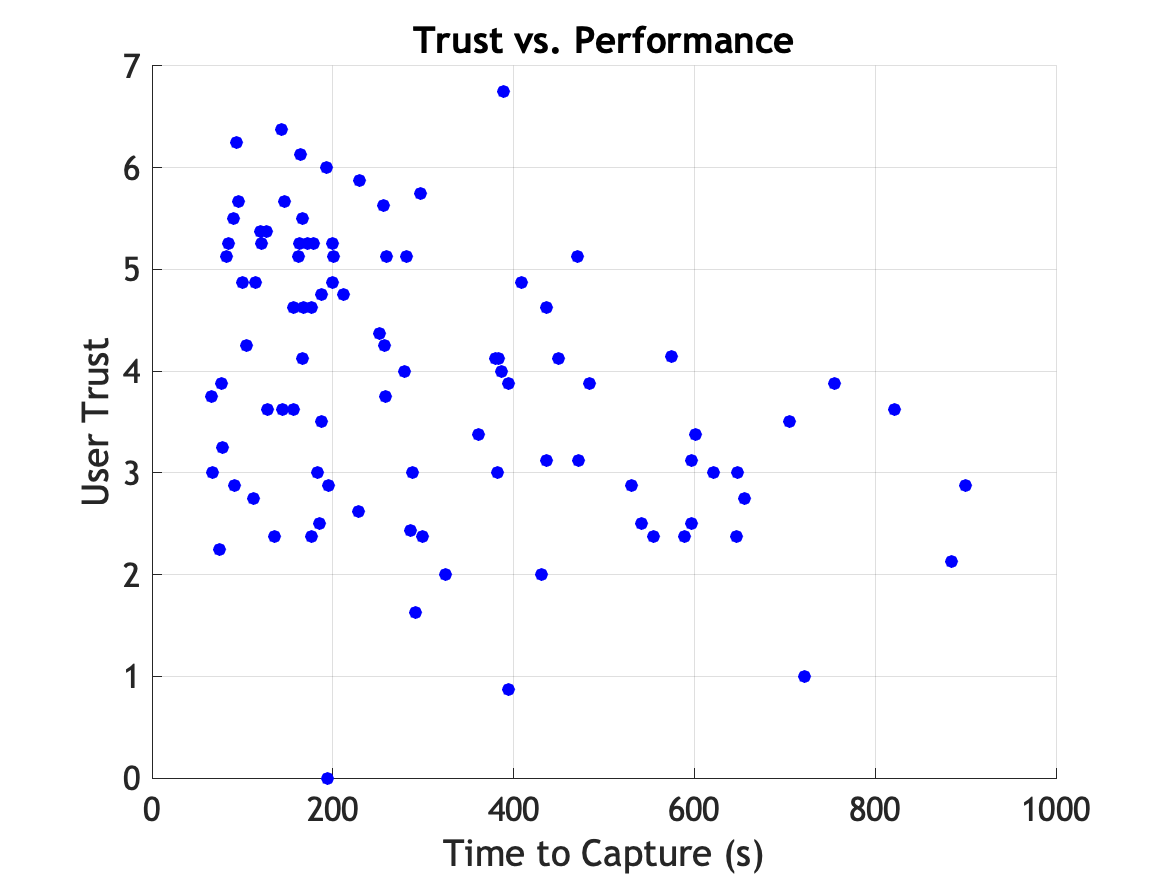}
	\caption{Correlation between user trust and performance across all \TROEdits{subject and mode} scenarios.}
    \label{fig:Human_Trust}
    \vspace{-12pt}
\end{figure}

\TROEdits{\textbf{H3} predicted that more effective users would lead to improved performance.} Semantic accuracies, $\eta$, for softmax and compass values were calculated according to Equations \ref{eq:softaccuracy} and \ref{eq:compassaccuracy}. When accuracies are evaluated for all scenarios, there are no clear correlations as capture times of ~200 seconds maintained a wide range of values from 0 (completely inaccurate) to 1 (completely accurate). This variance may be due to users providing helpful information that wasn't completely accurate but drew the drone towards where they may have seen to target. A more effective analysis compares each users' overall accuracy to their mean TTC for all three scenarios. These results are shown in Figure \ref{fig:Human_Accuracy}, which shows a statistically significant \TROEdits{($p<0.05$, $R^2=0.11$, $t(23)=-2.3$)} decreasing trend of lower user accuracy leading to poor performance. 

 However, user availability did not show any clear trends when compared to user performance. The average user availability was \TROEdits{57\%$\pm 19\%$}, which reflects the fact that users would answer about every other question posed by the robot. The frequency of questions asked by the robot varied throughout the simulation. The unpredictable nature caused some to users comment that it was challenging to notice that a question was being asked and respond to it in a prompt and accurate manner. Wide variability existed even for successful users as one subject captured the target in 188 seconds with an availability of 31\%, while another user finished at 168 seconds with an availability of 80\%. This observed contrast in conjunction with the accuracy trends demonstrates that the quality of information provided is more valuable than the quantity of observations. Hypothesis \textbf{H3} is therefore partially proven with the caveat that effective users are defined by higher accuracy, while availability is not a significant factor.


\TROEdits{\textbf{H4} predicted that situational awareness could be measured by camera view and lead to higher performance. However, the trends are} inconclusive. Users who observed more cameras per second performed slightly better than others, however not enough data is available to prove that this is indicative of a larger trend. There was also no relationship between the view rate and user accuracy. These results indicate that hypothesis \textbf{H4} cannot be substantiated.

\subsubsection{Subjective Metrics}
\TROEdits{\textbf{H5} predicted that more interaction with the robot would lead to higher operator workload. }Table \ref{tab:Performance} shows the total workload across the three different user modes. The average total reported workload across all three conditions demonstrates minor increases for \TROEdits{active ($\mu=3.2\pm0.15$), passive ($\mu=3.3\pm0.14$), and active/passive ($\mu=3.4\pm0.16$) modalities}. Overall, the workload did not seem to be significant with one user reporting that, \say{There was not a lot of workload}. Methods of reporting workload in a consistent manner across different subjects is an ongoing area of study \cite{TLXraw}. These challenges merit the consideration each subject’s relative ranking of the workload for each mode. Evaluating the relative workload from this perspective, the combined active/passive condition resulted in the highest level workload for 20/36 subjects, versus the active and passive cases respectively providing the highest workload for 16/36 subjects. \TROEdits{Therefore, while hypothesis \textbf{H5} cannot be substantiated it's worthy of note that the active/passive condition had the highest relative level of workload, whereas the passive and active scenarios displayed relatively equal rankings across subjects.}


\TROEdits{\textbf{H6} predicted that more interaction with the robot would lead to higher levels of trust. Evaluating the total reported trust across the different scenarios and subjects using a z-test, there is a moderately significant ($p<0.07$) difference between the reported trust for the active/passive case ($\mu=4.03\pm0.22$) versus the passive case ($\mu=3.47\pm0.24$), which had the lowest overall trust.} In post simulation questionnaires, one subject noted that \say{Having the drone ask questions and respond according to my answers built trust}. This may be indicative that giving the agent the capability to ask questions provides the user insight into its understanding of the mission. In evaluating hypothesis \textbf{H6}, there are no clear trends between user trust and user accuracy. However, in Figure \ref{fig:Human_Trust} a statistically significant \TROEdits{($p<0.01$, $R^2=0.15$, $t(93)=-4.2$)} trend developed between user trust and scenario mean TTC, with lower TTC strongly correlating to higher trust. 



\subsection{Discussion}
The human subject studies demonstrate that the proposed algorithm maintains strong performance while reasoning over a wide range observation accuracies and user behaviors from the provided semantic sketch information and query responses. The interaction did not significantly burden the users and the ability of the robot to offer bi-directional communication improved users' trust in the agent. Additional training may also improve the frustration felt by several users by clarifying robot intent and behavior. 

\TROEdits{The human subject studies also showed differences from the simulated study results. Most notable, the simulated results captured the targets more often than the humans did in the active case. This is likely due to the randomized starting conditions used in the simulated experiments versus the three static start points in human experiments. Furthermore, aspects of human cognition, in addition to accuracy and availability, that were unmodeled by the POMDP may have played a role. Future work will be required to identify additional cognitive features to improve cooperation with the human.}

Calculating the user’s accuracy as a single number disposes of the nuance in each observation. For example, the interaction in Figure \ref{fig:GoodUserBoth} shows the user providing an objectively incorrect first observation that the target was inside sketch \say{area1}. However, this information compelled the robot to move in that general direction and follow up questions to the user narrowed the robot’s belief of the target resulting in a capture. In this interaction, the user likely confused the proper location of the target on the 2D map, favoring a prompt coarse observation in the target’s general vicinity. 

In a different scenario (not shown), the user knowingly provided false information in order to steer the drone away from the town at the NW corner of the map. Due to the density of roads in this area, the drone often gravitated in this direction and user observations were helpful in getting it to explore other areas. In this case, a negative observation that the target was not in the town would have been equally helpful. Once out of the area, a subsequently accurate observation by the user helped the robot catch the target.

Despite these nuances, the significant trend correlating user’s overall softmax accuracy to performance, in comparison to the cruder compass metric, validates the implementation of the softmax approximation in Section IV.

\begin{figure}[h!]
    \centering
         \includegraphics[width=0.46\textwidth]{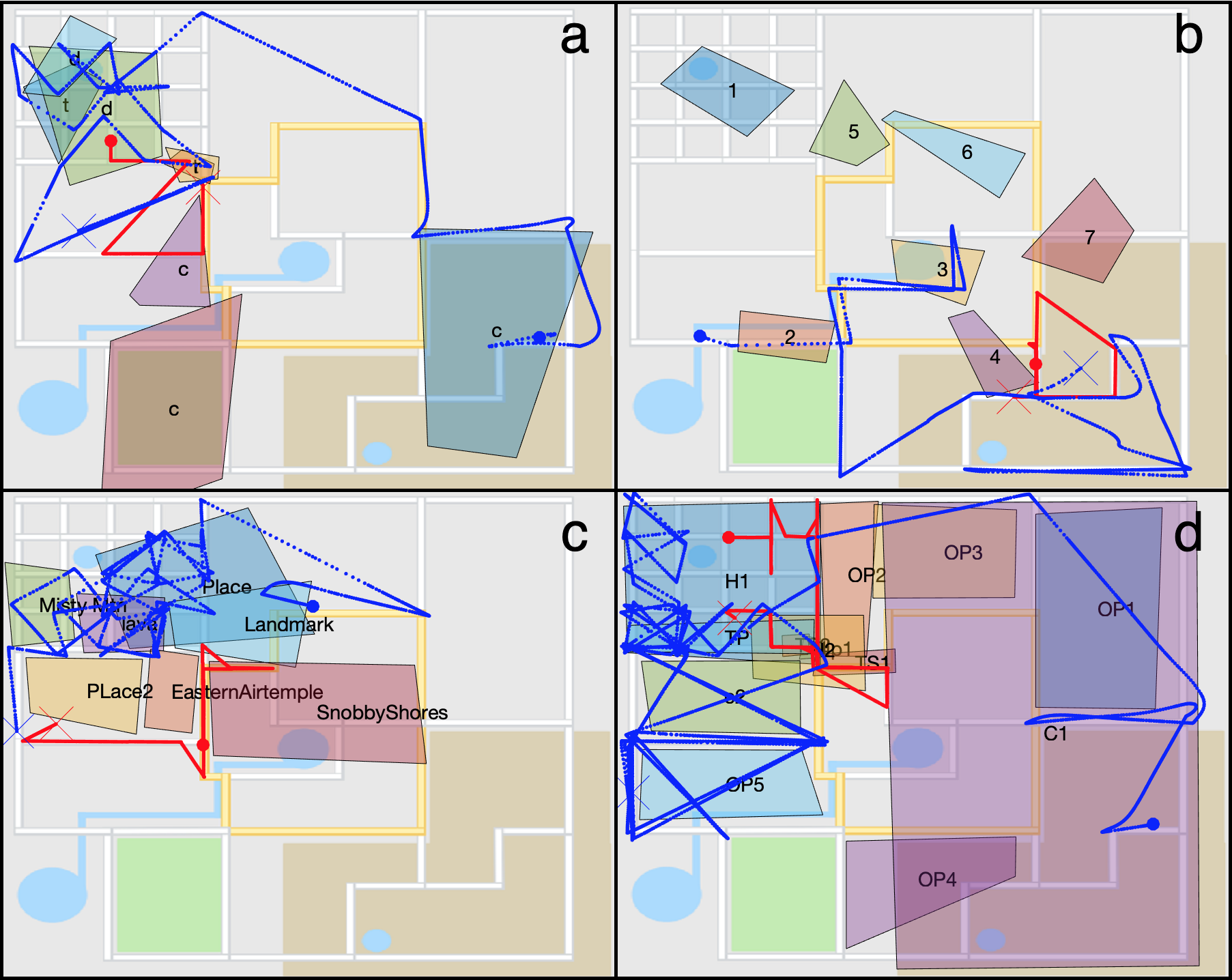}
         \caption{A collection of user sketches and interactions. In these interactions user \textbf{a} captured the target in 436s in the passive mode, user \textbf{b} captured the target in 288s in the active mode, user \textbf{c} captured the target in 575s in the active/passive mode, and user \textbf{d} failed to capture the target in the active/passive mode likely due to the significant number of sketches resulting in poor planning.}
         \label{fig:CuriousSketches}
\end{figure}

Notably, the algorithm encouraged users to interact with the interface in a personalized manner. Figure \ref{fig:CuriousSketches} shows how users chose to draw sketches around notable landmarks, at camera locations, or solely around the perceived target. Each of these methods resulted in the target being captured, particularly if the observations provided were followed up from numerous reference points. Users that failed to perform adequately often did not treat the robot as a teammate, instead trying to direct the drone to regions that they wanted it to visit, such as in Figure \ref{fig:BadUserPush}, or giving a standalone observation from a single reference point. While standalone observations were considered sufficient by some users, the robot’s ability to ask follow-up questions showed an improvement in performance and a significant increase in user trust. However, numerous subjects reported frustration in dealing with the robot as it seemed to not take their observations into account. Due to the non-intuitive nature of the actions determined by the online POMDP planner, the drone displayed complex behavior such as not going directly to the area mentioned to cut-off possible escape locations. Sometimes, this frustration turned to surprise as the target was subsequently caught, such as when one user noted, \say{I still couldn't figure out what the robot was doing sometimes and why it was flying in certain directions. This made me think it wasn't as consistent and predictable. It was quite competent in the end at accomplishing the task.} Subjects mentioned that continued interactions with the robot improved their understanding of effective interaction methods commenting that, \say{I think the practice from the previous rounds made it easier to determine what to do this round.} In our study, the subject briefing was left intentionally vague to evaluate the intuitive nature of the interaction. However, like any collaborative task with a teammate, joint training would likely improve performance as robot behavior becomes less opaque.

To improve the collaborative nature of the searching task, future work may incorporate non-binary observations of the target. For example, instead of only being offered an option of \say{Is/Is not} in the structured language observation menu, a user could express their own uncertainty with nuanced observations such as probably/probably not. While these observations would be more challenging for the robot to reason through, the user may be compelled to provide extra information when they have only a vague understanding of the target location. Adding degrees of uncertainty to observations naturally extends the uncertainty enabled through the sketch-based interface and replicates how humans communicate to each other in dynamic environments.


\section{Conclusion}
This paper presented and demonstrated the HARPS framework, a novel approach to multi-level active semantic sensing and planning in human-robot teams in unknown dynamic environments. We extend online POMDP planning frameworks to incorporate semantic soft data from a human sensor about uncertain dynamic task states, 
shared from human to robot through the use of a sketch-based semantic structured language interface. Enabling this interaction is a novel formulation of a POMDP with a ``human-in-the-loop" active sensing model, as well as innovations to the use of soft-data fusion that allow the communication of higher level modal information. The approach was demonstrated and evaluated on a relevant dynamic target search problem with real and simulated operators, which highlighted the improvements that queries to a human can bring to a collaborative target search problem. The results show the simulated robot effectively using localized sketches and binary observations in response to language-based queries to intercept a moving target. The human subject studies validated the effectiveness of the algorithm in extracting information from users in a wide range of interaction styles, and showed that active querying increased user trust in the robot without adding excessive workload. 

The realized system enables robust, intuitive human-robot collaboration in unknown, dynamic environments. The introduction of a dialog based on sketches and colloquial language creates new opportunities for operators to engage with robotic systems in diverse manners, enabling the communication of accurate and useful information without the need to insist on machine level precision from humans. Furthermore, extensive live and simulated testing inspires high confidence in the system to adapt to any human, rather than relying on operators with specific training, skill sets, or knowledge bases. This combination of decreased reliance on high precision humans and increased tolerance of human-introduced uncertainty dramatically expands the space of useful applications for human-robot teaming with respect to the previous state of the art.


\bibliographystyle{IEEEtran}
\bibliography{refs}


\begin{IEEEbiography}[{\includegraphics[width=1in,height=1.5in,clip,keepaspectratio]{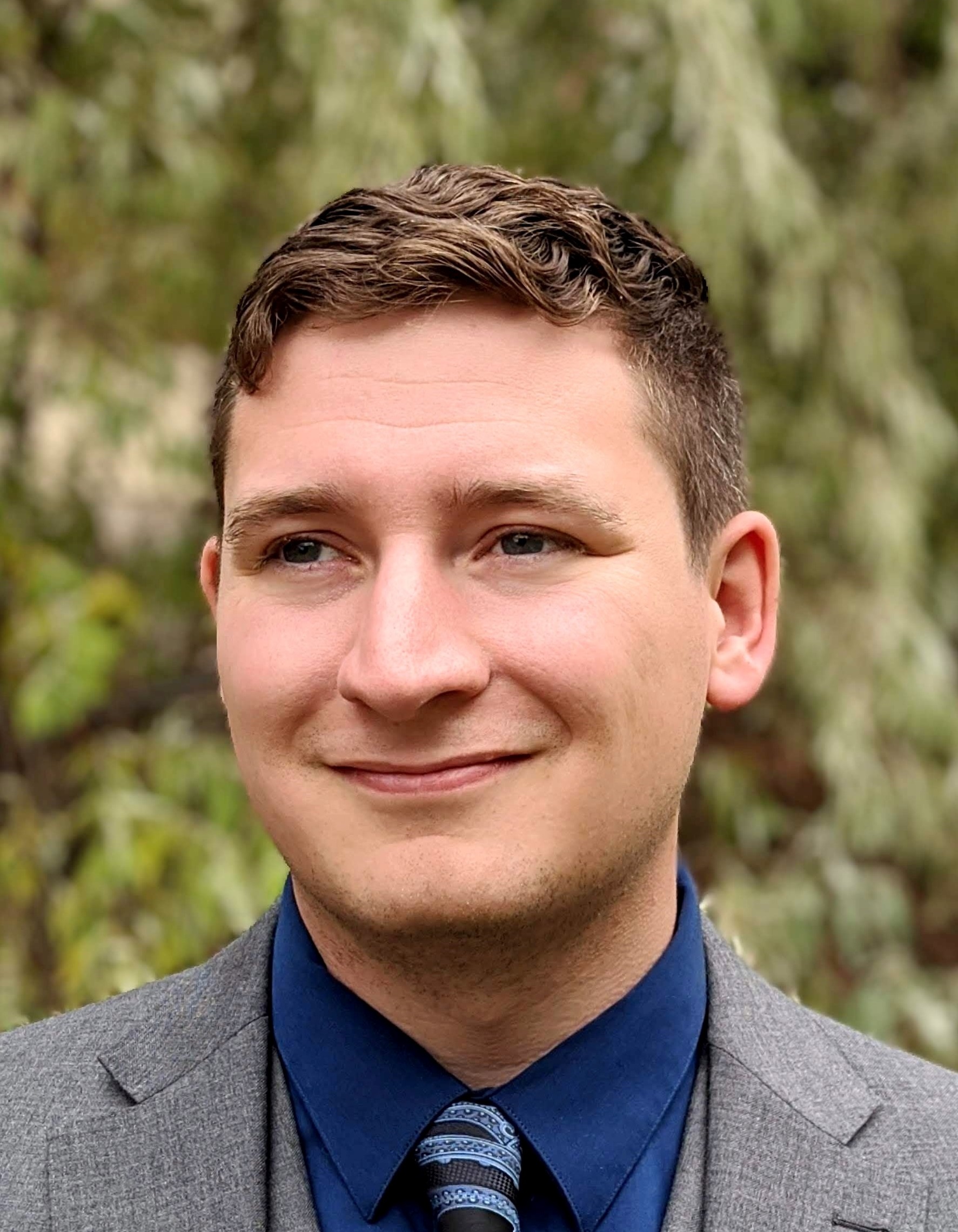}}]{Luke Burks}
is currently a Research Engineer and co-lead of the Humans and Autonomy Research Group at Aurora Flight Sciences. He received a B.S in Physics from the University of Arkansas in 2015, and his Ph.D. in Aerospace Engineering Sciences from the University of Colorado Boulder in 2020, focusing on optimal planning and sensing in human-robot teams. 

His research focuses on algorithms for optimal decision making under uncertainty and probabilistic inference for human-autonomy interaction. 
\end{IEEEbiography}

\begin{IEEEbiography}[{\includegraphics[width=1in,height=1.5in,clip,keepaspectratio]{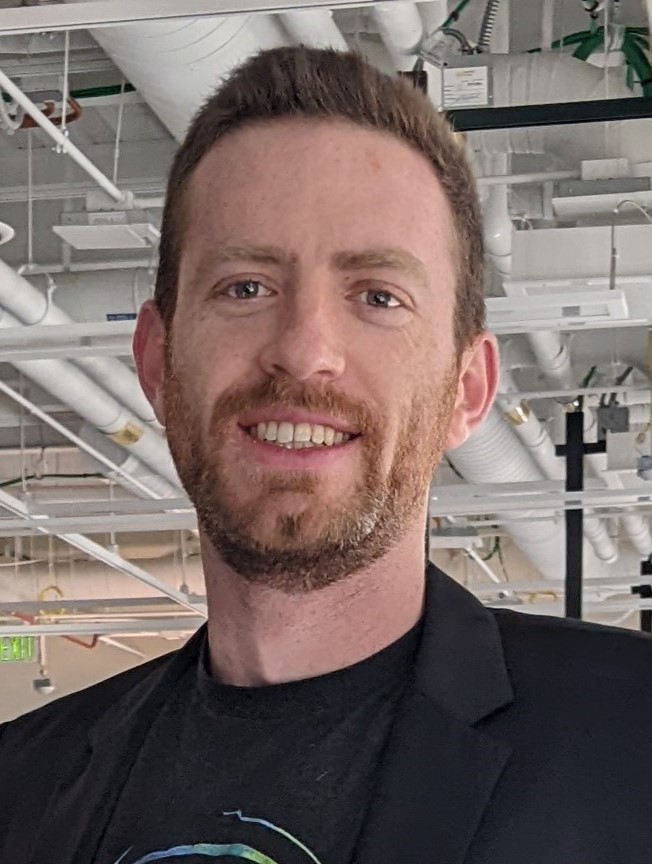}}]{Hunter M. Ray}
received his B.Sc in Mechanical Engineering from Brown University, Providence, RI, USA in 2018 and his M.S in Aerospace Engineering from the University of Colorado Boulder, Boulder, CO, USA in 2021. He is currently working towards his Ph.D. in aerospace engineering from the Ann and H.J. Smead Aerospace Engineering Sciences Department at the University of Colorado Boulder.

His research interests include the development of intuitive robotic interfaces and probabilistic algorithms that enable human-autonomy teams.
\end{IEEEbiography}

\begin{IEEEbiographynophoto}{Jamison McGinley}
received his B.S. and M.S degree in aerospace engineering from the University of Colorado Boulder, Boulder, CO, USA, respectively in 2019 and 2022.
\end{IEEEbiographynophoto}

\begin{IEEEbiographynophoto}{Sousheel Vunnam}
received his B.S. degree in computer science from the University of Colorado Boulder, Boulder, CO, USA in 2019.
\end{IEEEbiographynophoto}

\begin{IEEEbiography}[{\includegraphics[width=1in,height=1.5in,clip,keepaspectratio]{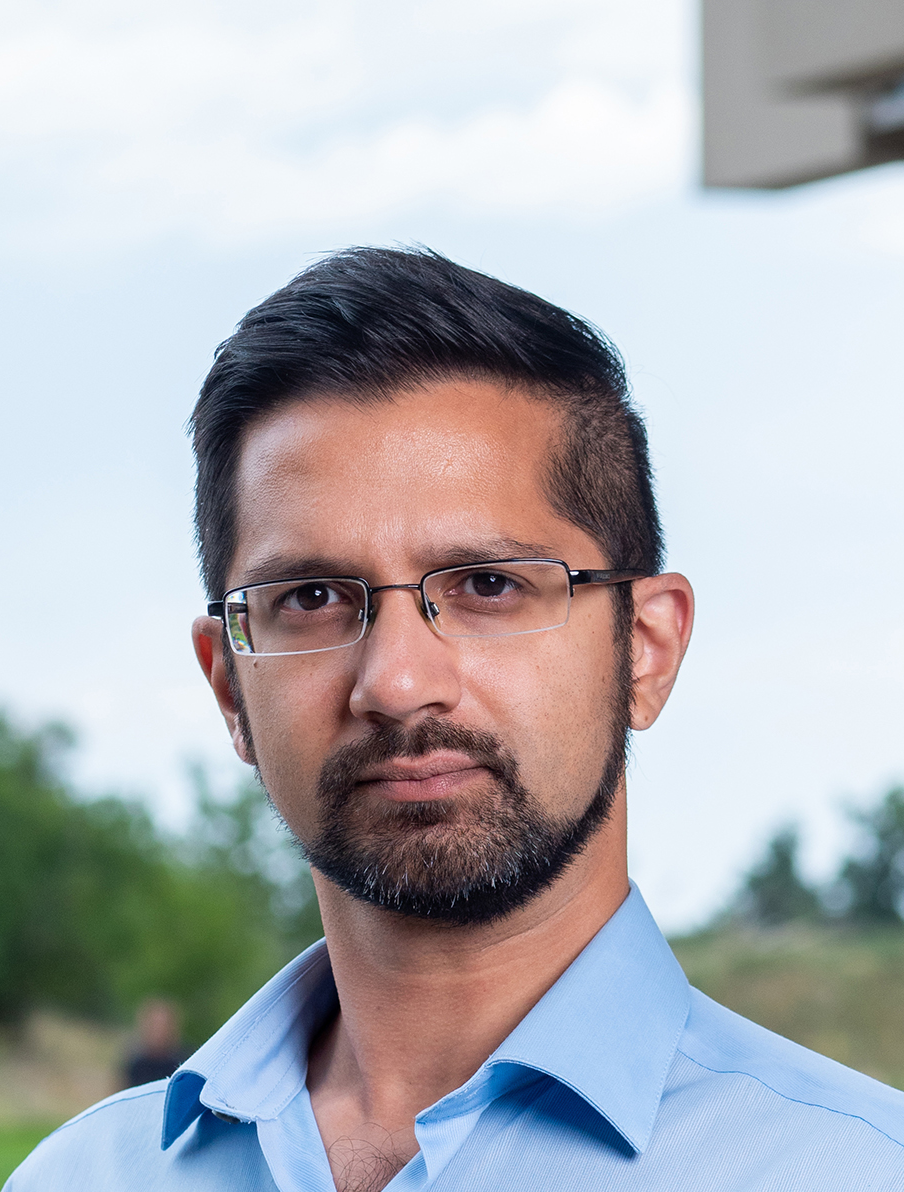}}]{Nisar Ahmed}
is an Associate Professor and H.J. Smead Faculty Fellow in the Smead Aerospace Engineering Sciences Department at the University of Colorado Boulder. He earned his B.S. in Engineering in 2006 from Cooper Union (New York, NY) and M.S. and Ph.D. in Mechanical Engineering in 2012 from Cornell University (Ithaca, NY). 
He 
directs the Cooperative Human-Robot Intelligence (COHRINT) Lab at the University of Colorado Boulder, which researches probabilistic modeling, estimation and control of autonomous systems, human-robot/machine interaction, sensor fusion, and decision-making under uncertainty. 
\end{IEEEbiography}



\end{document}